\documentclass[10pt,twocolumn,letterpaper]{article}
\usepackage[pagenumbers]{cvpr} % To force page numbers, e.g. for an arXiv version
\usepackage{graphicx}
\usepackage{amsmath}
\usepackage{amssymb}
\usepackage{booktabs}
\usepackage{multirow}
\usepackage{caption}
\usepackage{subcaption}

\usepackage{amsmath,amsfonts,bm}

\def\eqref#1{equation~(\ref{#1})}
\def\1{\bm{1}}

\def\vzero{{\bm{0}}}

\def\vb{{\bm{b}}}

\def\vl{{\bm{l}}}

\def\vo{{\bm{o}}}

\def\vw{{\bm{w}}}
\def\vx{{\bm{x}}}

\def\mQ{{\bm{Q}}}
\def\mR{{\bm{R}}}

\def\mW{{\bm{W}}}
\def\mX{{\bm{X}}}

\def\mLambda{{\bm{\Lambda}}}

\DeclareMathAlphabet{\mathsfit}{\encodingdefault}{\sfdefault}{m}{sl}
\SetMathAlphabet{\mathsfit}{bold}{\encodingdefault}{\sfdefault}{bx}{n}

\newcommand{\R}{\mathbb{R}}

\newcommand{\softmax}{\mathrm{softmax}}

\def\versus{\emph{vs}\onedot} 
\usepackage[pagebackref,breaklinks,colorlinks]{hyperref}
\usepackage[capitalize]{cleveref}
\crefname{section}{Sec.}{Secs.}
\Crefname{section}{Section}{Sections}
\Crefname{table}{Table}{Tables}
\crefname{table}{Tab.}{Tabs.}
\def\cvprPaperID{5401} % *** Enter the CVPR Paper ID here
\def\confName{CVPR}
\def\confYear{2022}

\begin{document}
\title{ViM\@: Out-Of-Distribution with Virtual-logit Matching}

\author{Haoqi Wang\(^1\)\textsuperscript{\textasteriskcentered}\quad Zhizhong Li\(^1\)\thanks{~These two authors contribute equally to the work.}\textsuperscript{\textasteriskcentered}~\quad Litong Feng\(^1\)\quad Wayne Zhang\(^{12}\)\thanks{~Corresponding author: Wayne Zhang.}\\
\(^1\)SenseTime Research \quad \(^2\)Qing Yuan Research Institute, Shanghai Jiao Tong University
\\
{\tt\small \{wanghaoqi,lizz,fenglitong,wayne.zhang\}@sensetime.com}
}
\maketitle
\begin{abstract}
    Most of the existing Out-Of-Distribution (OOD) detection algorithms depend on single input source: the feature, the logit, or the softmax probability.
However, the immense diversity of the OOD examples makes such methods fragile.
There are OOD samples that are easy to identify in the feature space while hard to distinguish in the logit space and vice versa.
Motivated by this observation, we propose a novel OOD scoring method named Virtual-logit Matching (ViM),
which combines the class-agnostic score from feature space and the In-Distribution (ID) class-dependent logits.
Specifically, an additional logit representing the virtual OOD class is generated from the residual of the feature against the principal space,
and then matched with the original logits by a constant scaling.
The probability of this virtual logit after softmax is the indicator of OOD-ness.
To facilitate the evaluation of large-scale OOD detection in academia, we create a new OOD dataset for ImageNet-1K, which is human-annotated and is \(8.8\times\) the size of existing datasets.
We conducted extensive experiments, including CNNs and vision transformers, to demonstrate the effectiveness of the proposed ViM score.
In particular, using the BiT-S model, our method gets an average AUROC 90.91\% on four difficult OOD benchmarks, which is 4\% ahead of the best baseline.
Code and dataset are available at \url{https://github.com/haoqiwang/vim}.

\end{abstract}

\section{Introduction}\label{sec:intro}

Considering most deep image classification models are trained in the closed-world setting, the \emph{out-of-distribution} (OOD) issue arises and deteriorates customer experience when the models are deployed in production, facing inputs coming from the open world~\cite{openworld06esi}.
For instance, a model may wrongly but confidently classify an image of \emph{crab} into the \emph{clapping} class, even though no crab-related concepts appear in the training set.
OOD detection is to decide whether an input belongs to the training distribution.
OOD detection complements classification and finds its application in fields such as autonomous driving~\cite{kitt2010visual}, medical analysis~\cite{schlegl2017unsupervised} and industrial inspection~\cite{mvtec19cvpr}.
A comprehensive review of OOD and related topics including open set recognition, novelty detection and anomaly detection can be found in~\cite{yang2021generalized}.

\begin{figure}[t]
    \begin{center}
        \includegraphics[width=0.4\textwidth]{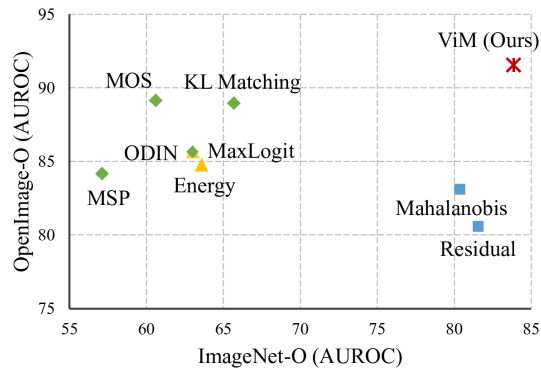}
        % \vspace{-0.4cm}
    \end{center}
    \caption{
        The AUROC (in percentage) of nine OOD detection algorithms applied to a BiT model trained on ImageNet-1K.
        The OOD datasets are ImageNet-O (\(x\)-axis) and OpenImage-O (\(y\)-axis).
        Methods marked with box \(\Box\) use the feature space;
        methods with triangle \(\triangle\) use the logit;
        and methods with diamond \(\Diamond\) use the softmax probability.
        The proposed method ViM (marked with *) uses information from both features and logits.
    }\label{fig:teaser}
    \vspace{-0.4cm}
\end{figure}

The core of an OOD detector is a scoring function \(\phi\) that maps an input feature \(\vx\) to a scalar in \(\R\), indicating to what extent the sample is likely to be OOD\@.
In testing, a threshold \(\tau\) is decided, ensuring that the validation set retains at least a given true-positive rate (TPR), \eg the typical value of \(0.95\).
The input example is regarded as OOD if \(\phi(\vx)>\tau\) and as ID (\ie, in-distribution) otherwise.
In cases where a score indicating the ID-ness is convenient,
we can mentally use the negative of OOD score as the ID score.

Researchers have designed quite a few scoring functions
by seeking properties that are naturally held by ID examples and easily violated by OOD examples, or \emph{vice versa}.
Scores are mainly derived from three sources:
(1) \emph{the probability}, such as
the maximum softmax probabilities~\cite{msp17iclr},
the minimum KL-divergence between the softmax and the mean class-conditional distributions~\cite{hendrycks2019scaling};
(2) \emph{the logit}, such as
the maximum logits~\cite{hendrycks2019scaling},
the \(\mathrm{logsumexp}\) function over logits~\cite{energyood20nips};
and (3) \emph{the feature}, such as
the norm of the residual between feature and the pre-image of its low-dimensional embedding~\cite{ndiour2020out},
the minimum Mahalanobis distance between the feature and the class centroids~\cite{mahananobis18nips}, \etc.
In these methods, OOD scores can be directly computed from existing models without re-training,
making the deployment effortless.
However, as illustrated in \cref{fig:teaser}, their performances are limited by the singleness of their information source:
using features exclusively disregards the classification weights with class-dependent information;
using the logit or the softmax solely misses feature variations in the null space~\cite{cook2020outlier}, which carries class-agnostic information;
and the softmax further discards the norm of logits.
To cope with the immense diversity that manifests in OOD samples,
we ask the question,
\emph{is it helpful to design an OOD score that utilizes multiple sources?}

Built upon the success of prior arts, we design a novel scoring function termed the \emph{Virtual-logit Matching}  (ViM) score,
which is the softmax score of a constructed virtual OOD class whose logit is jointly determined by the feature and the existing logits.
To be specific, the scoring function first extracts the residual of the feature against a principal subspace, and then converts it to a valid logit by matching its mean over training samples to the average maximum logits.
Finally, the softmax probability of the devised OOD class is the OOD score.
From the construction of ViM, we can see intuitively that the smaller the original logits and the greater the residual, the more likely it is to be OOD\@.

Different from the aforementioned methods, another line of works tailors the features learned by the network to better identify ID and OOD by imposing dedicated regularization losses~\cite{confbranch2018arxiv,godin20cvpr,huang2021mos,onedim21cvpr} or by exposing generated or real collected OOD samples~\cite{yang2021scood,confcal18iclr}.
As they all require the re-training of the network, we briefly mention them here and will not delve into the details.

Recently, OOD detection in large-scale semantic space has attracted increasing attention~\cite{roady2019out,hendrycks2019scaling,hendrycks2021natural,huang2021mos}, advancing OOD detection methods toward real-world applications.
However, the current shortage of clean and realistic OOD datasets for large-scale ID datasets becomes an impediment to the field.
Previous OOD datasets were curated from public datasets which were collected with a predefined tag list, such as iNaturalist, Texture, and ImageNet-21k (\cref{tab:datasets}).
This may lead to a biased performance comparison, specifically, the hackability of small coverage as described in \cref{sec:openimage}.
To avoid this risk, we build a new OOD benchmark for ImageNet-1K~\cite{deng2009imagenet} models, \emph{OpenImage-O}, from OpenImage dataset~\cite{openimages} with natural class distribution.
It contains 17,632 manually filtered images, and is \(7.8\times\) larger than the recent ImageNet-O~\cite{hendrycks2021natural} dataset.

\begin{table}[t]
    \centering
    \footnotesize
    \setlength\tabcolsep{2 pt}
    \begin{tabular}{llrr}
        \toprule
        \textbf{Dataset}                                   & \textbf{Image Distribution} & \textbf{\#Image} & \textbf{Labeling Method} \\
        \midrule
        OpenImage-O                                        & natural class statistics    & \(17,632\)       & image-level manual       \\
        Texture~\cite{cimpoi14describing}                  & predefined tag list         & \(5,160\)        & tag-level manual         \\
        iNaturalist~\cite{van2018inaturalist,huang2021mos} & predefined tag list         & \(10,000\)       & tag-level manual         \\
        ImageNet-O~\cite{huang2021mos}                     & hard adversarial OOD        & \(2,000\)        & image-level manual       \\
        \bottomrule
    \end{tabular}
    \caption{
        OpenImage-O follows natural class statistics, while ImageNet-O is adversarially built to be hard.
        Both datasets have image-level OOD annotation.
        Texture and iNaturalist are selected by tags, and their OOD labels are annotated in tag-level.
    }\label{tab:datasets}
\end{table}

We extensively evaluate our method on various models using ImageNet-1K as the ID dataset.
The model architectures range from the classical ResNet-50~\cite{he2019bag}, to the recent BiT~\cite{kolesnikov2020big}, and to the latest ViT-B16~\cite{dosovitskiy2021an}, RepVGG~\cite{ding2021repvgg}, DeiT~\cite{pmlr111} and Swin Transformer~\cite{liu2021swin}.
From the results on four OOD datasets, including OpenImage-O, ImageNet-O, Texture, and iNaturalist,
we found that model selection affected the performance of many baseline methods, while our method performs stably well.
Specially, our method achieved an average AUROC of 90.91\% using the BiT model, which greatly surpasses the best baseline whose average AUROC is 86.62\%.

Our contributions are threefold.
(1) We proposed a novel OOD detection method ViM, that works well for a large range of models and datasets, owing to the effective fusion of information from both features and logits.
The method is lightweight and fast, requiring neither extra OOD data nor re-training.
(2) We conducted comprehensive experiments and ablation studies on the ImageNet-1K dataset, including CNNs and vision transformers.
(3) We curated a new OOD dataset for ImageNet-1K called OpenImage-O, which is very diverse and contains complex scenes.
We believe it will facilitate research on large-scale OOD detection.

\section{Related Work}\label{sec:related}

\paragraph{OOD/ID Score Design}

Hendrycks \etal~\cite{msp17iclr} presented a baseline method using the maximum predicted softmax probability (MSP) as the ID score.
ODIN~\cite{odin18iclr} enhances MSP by perturbing the inputs and rescaling the logits.
Hendrycks \etal~\cite{hendrycks2019scaling} also experimented with the MaxLogit and the KL matching method on the ImageNet dataset.
The energy score~\cite{energyood20nips} computes the \(\mathrm{logsumexp}\) on logits,
and ReAct~\cite{sun2021tone} strengthens the energy score by feature clipping.
In~\cite{ndiour2020out} the norm of the difference between the feature and the pre-image of its low-dimensional manifold embedding is used.
Lee \etal~\cite{mahananobis18nips} computes the minimum Mahalanobis distance between the feature and the class-wise centroids.
NuSA~\cite{cook2020outlier} uses the ratio of the norm of feature projected onto the column space of the classification weight matrix to the original norm as the ID score.
The gradients are also used as evidence for ID and OOD distinction in~\cite{huang2021importance}.
For methods using logits/probabilities,
feature variations on the null space of the weight matrix are completely ignored;
while for methods that operate on the features space,
the class-dependent information on weight matrix is dropped.
Our method combines the strengths of feature-based scores and logit-based scores by the novel mechanism of virtual logit,
and gets substantial improvements.

\paragraph{Network/Loss Design}

Many works redesign the training loss to be OOD-aware~\cite{confbranch2018arxiv} or add regularization terms~\cite{onedim21cvpr,huang2021mos} to push part ID/OOD features.
DeVries \etal~\cite{confbranch2018arxiv} augment the network with a confidence estimation branch that uses misclassified in-distribution examples as a proxy for out-of-distribution examples.
MOS~\cite{huang2021mos} modifies the loss to use the pre-defined group structure so that the minimum group-wise ``else'' class probability can indicate the OOD-ness.
Zaeemzadeh \etal~\cite{onedim21cvpr} forces the ID samples to embed into a union of 1-dimensional subspaces during training and computes the minimum angular distance from the feature to the class-wise subspaces.
Generalized ODIN~\cite{godin20cvpr} uses a dividend/divisor structure to encode the prior knowledge of decomposing the confidence of class probability.
Different from these methods, our method does not require model retraining, thus not only is it easier to apply, but the ID classification accuracy is also preserved.

\paragraph{OOD Data Exposure}
Outlier Exposure~\cite{oe18nips} utilizes an auxiliary OOD dataset to improve OOD detection.
Dhamija \etal~\cite{agnostophobia18nips} regularize samples from extra background classes to have uniform logits and to have small feature norms.
Lee \etal~\cite{confcal18iclr} use GAN to generate OOD samples that lie near the ID samples and push the prediction of OOD samples to the uniform distribution.
Several methods, including MCD~\cite{mcd19iccv}, NGC~\cite{ngc21iccv} and UDG~\cite{yang2021scood}, can utilize external unlabeled noisy data to enhance the OOD detection performances.
Different from these methods, our method does not require additional OOD data and thus avoids biases towards the introduced OOD samples~\cite{lessbias19bmvc}.

\section{Motivation: The Missing Info in Logits}\label{sec:motivation}

\begin{figure}[t]
    \begin{center}
        \includegraphics[width=0.45\textwidth]{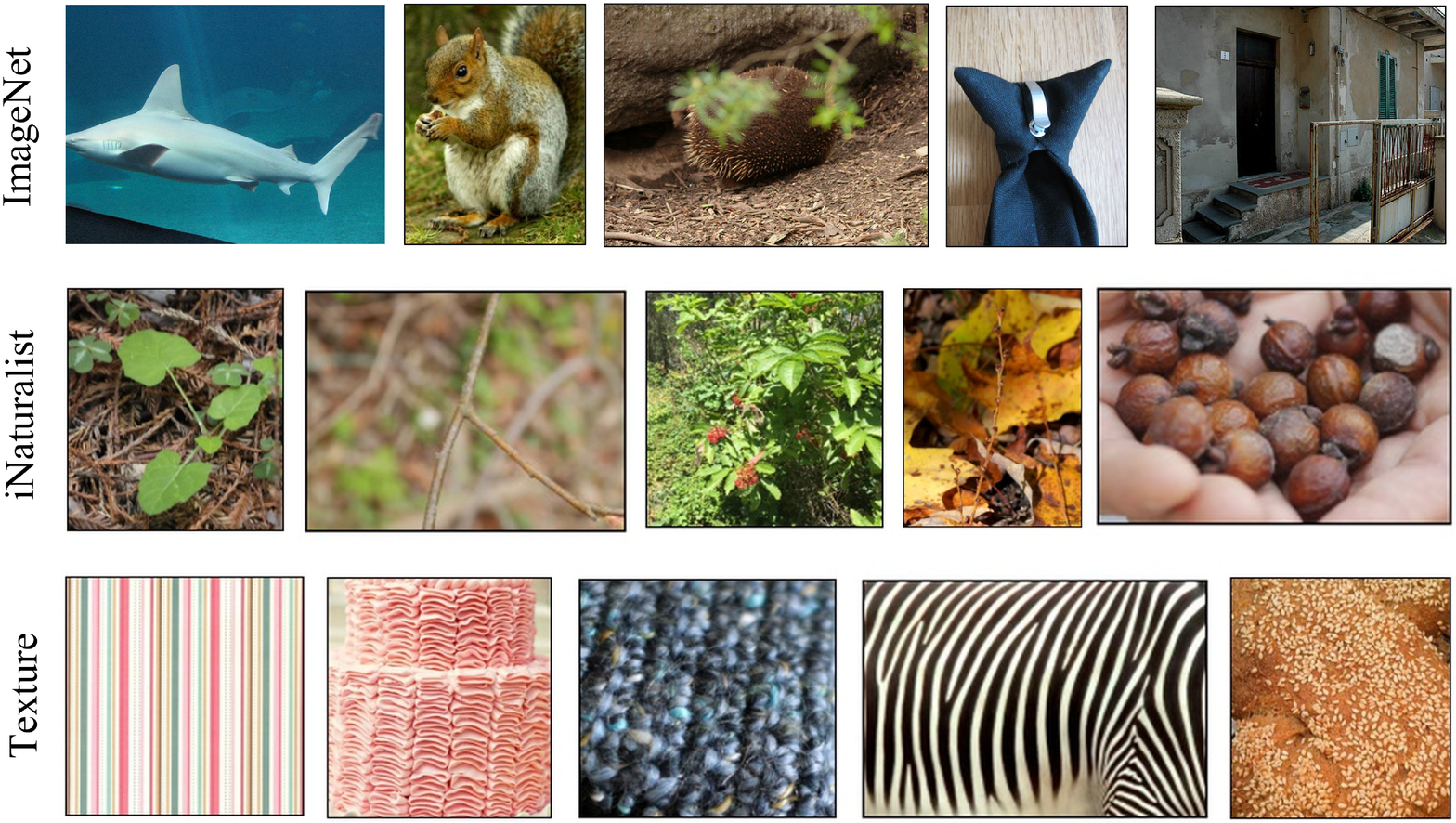} \\
        \vspace{1.2em}
        \includegraphics[width=0.35\textwidth]{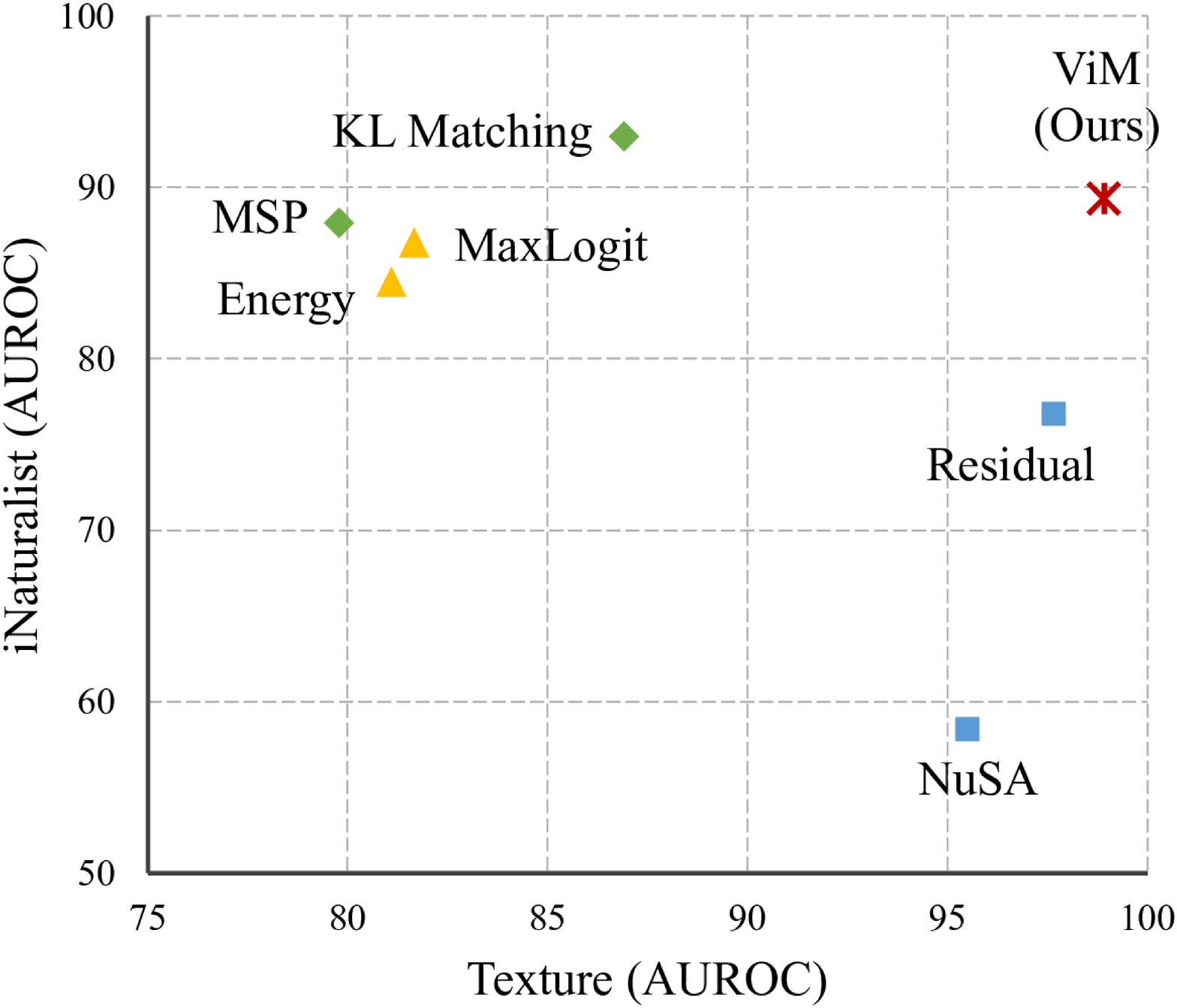}
        \vspace{-0.4cm}
    \end{center}
    \caption{
        Comparison of AUROC for OOD detection algorithms that are based on probability (marked with diamond \(\Diamond\)), logit (\(\triangle\)), and feature (\(\Box\))
        of 9 OOD detection algorithms applied to a BiT model trained on ImageNet-1K.
        The OOD datasets are Texture (\(x\)-axis) and iNaturalist (\(y\)-axis).
        Example images for the ID dataset ImageNet-1K and the two OOD datasets are illustrated at the top.
    }\label{fig:hack_dataset}
    \vspace{-0.4cm}
\end{figure}

For a series of OOD detection methods that are based on logits or softmax probabilities, we find that their performances are limited.
In~\cref{fig:teaser}, feature-based OOD scores such as Mahalanobis and Residual are good at detecting OOD in ImageNet-O, while all methods that are based on logit/probability lag behind.
This is not an accident, as is again shown in~\cref{fig:hack_dataset}.
The AUROC of the state-of-the-art probability-based method KL Matching is still lower than straightforwardly designed OOD scores in feature space on Texture dataset.
This motivates us to study the influence of the lost information going from features to logits.

Consider a \(C\)-class classification model whose logit \(\vl\in\R^C\) is transformed from the feature \(\vx\in\R^N\) by a fully connected layer with weight \(\mW\in\R^{N\times C}\) and bias \(\vb\in\R^C\), \ie \(\vl = \mW^T \vx + \vb\).
The predicted probability is \(p(\vx) = \softmax(\vl)\).
For convenience, we set the point \(\vo:=-(\mW^T)^+\vb\),
where \((\cdot)^+\) is the Moore-Penrose inverse,
as the origin of a new coordinate system of feature space,
\begin{equation}\label{eq:bias-free}
    \vl = \mW^T \vx' = \mW^T (\vx - \vo), \quad \forall \vx.
\end{equation}
Geometrically,
each logit \(l_i\) is the inner product between the feature \(\vx'\) and the class vector \(\vw_i\) (the \(i\)-th column of \(\mW\)).
Later when generalizing logits to virtual logits, we will replace \(\vw_i\) with a subspace, and replace the inner product with a projection.
The bias term is safely omitted in the new coordinate system.
In the remaining part of the paper, we assume the feature space uses the new coordinate system.
Logits contain class-dependent information,
yet there is class-agnostic information in feature space that is not recoverable from logits.
We study two cases (null space and principal space) and discuss the two OOD scores (NuSA and Residual) that rely on them, respectively.

\paragraph{OOD Score Based on Null Space}
A feature \(\vx\) can be decomposed into \(\vx = \vx^{W^\perp} + \vx^W\),
where \(W\) is the column space of \(\mW\),
\(\vx^{W^\perp}\) and \(\vx^W\) are projections of \(\vx\) to \(W^\perp\) and \(W\), respectively.
\(W^\perp\) is the null space of \(\mW^T\), and we have \(\mW^T\vx^{W^\perp}=\vzero\).
The component \(\vx^{W^\perp}\) does not affect classification, but it influences OOD detection.
It is demonstrated in~\cite{cook2020outlier} that one can perturb an image intensely yet constrain the difference between the features in \(W^\perp\).
The resulting outlier images are not like any of the ID images but retains high confidence in classification.
Taking advantage of this, they define an ID score NuSA (null space analysis) as
\begin{equation}
    {\text{NuSA}}(\vx) = \frac{\sqrt{\|\vx\|^2-\|\vx^{W^\perp}\|^2}}{\|\vx\|}.
\end{equation}
Intuitively, NuSA uses the angle (\(=\arccos({\text{NuSA}}(\vx))\)) between \(\vx\) and \(W\) to indicate the OOD-ness.
From \cref{fig:hack_dataset} we can see that the simple angle information clearly distinguishes OOD examples in Texture with an AUROC 95.50\%, surpassing methods based on logits and the competitive method KL Matching based on softmax probability.

\paragraph{OOD Score Based on Principal Space}

It is generally assumed that features lie in low-dimensional manifolds~\cite{ndiour2020out,onedim21cvpr}.
For simplicity, we use linear subspace (in the new coordinate system) passing through the origin \(\vo\) as the model.
We define the \emph{principal space} as the \(D\)-dimensional subspace \(P\) spanned by eigenvectors of the largest \(D\) eigenvalues of the matrix \(\mX^T \mX\), where \(\mX\) is the ID data matrix.
Features that deviate from the principal space are likely to be OOD examples.
We can define
\begin{align}
    {\text{Residual}}(\vx) & = \|\vx^{P^\perp}\|,\label{eq:residual}
\end{align}
to capture the deviation of features from the principal space.
Here \(\vx = \vx^P+\vx^{P^\perp}\) and \(\vx^{P^\perp}\) is the projection of \(\vx\) to \({P^\perp}\).
The residual score is similar to the reconstruction error in~\cite{ndiour2020out} except that they employ nonlinear manifold learning for dimension reduction.
Note that after the projection onto logits, this deviation is corrupted since the matrix \(\mW^T\) projects to a lower dimensional space than the feature space.
\cref{fig:hack_dataset} shows that Residual score improves over the NuSA score on both datasets,
making the performance contrast between feature-based methods with logit/probability-based methods more striking.

\paragraph{Fusing Class-dependent and Class-agnostic Information}
In contrast to methods on logit/probability, both the NuSA and the Residual do not consider information that is specific to individual ID classes, namely they are class-agnostic.
As a consequence, these scores ignore the feature similarity to each ID class, and are ignorant about which class the input resembles most.
This gives an explanation of their worse performance on the iNaturalist OOD benchmark, as iNaturalist samples need to distinguish subtle differences between fine-grained classes.
We hypothesize that unifying the information from feature space and the logits could improve the detection performance on a broader type of OOD samples.
Such a solution is presented in \cref{sec:method} using the concept of virtual logit.

\begin{figure*}[t]
    \begin{center}
        \includegraphics[width=0.85\textwidth]{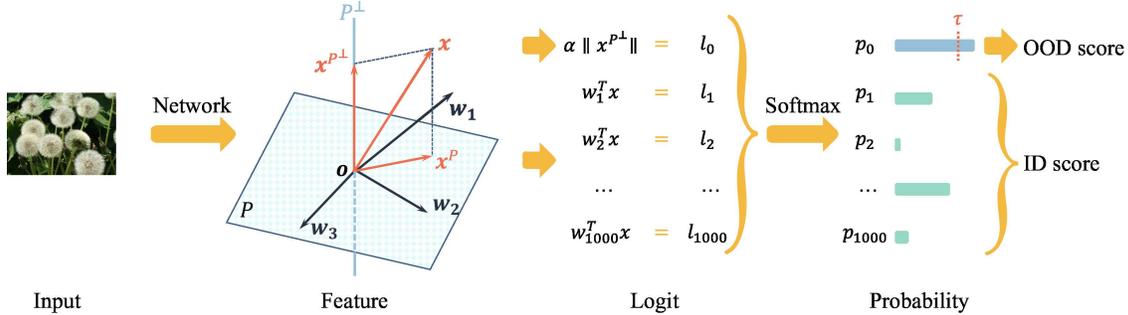}
    \end{center}
    \caption{
        The pipeline of ViM\@.
        The principal space \(P\) and the matching constant \(\alpha\) are determined by the training set beforehand using \cref{eq:eigen} and \cref{eq:alpha}.
        In inference, feature \(\vx\) is computed by the network, and the virtual logit \(\alpha\|\vx^{P^\perp}\|\) is computed by projection and scaling.
        After softmax, the probability corresponding to the virtual logit is the OOD score.
        It is OOD if the score is larger than threshold \(\tau\).
    }\label{fig:pipeline}
\end{figure*}

\section{Virtual-logit Matching}\label{sec:method}

To unify the class-agnostic and class-dependent information for OOD detection, we propose an OOD score by Virtual-logit Matching, abbreviated as ViM\@.
The pipeline is illustrated in \cref{fig:pipeline}, where there are three steps, operating at the feature, the logit, and the probability, respectively.
To be specific,
for feature \(\vx\),
(1) extract the residual \(\vx^{P^\perp}\) of \(\vx\) against the principal subspace \(P\);
(2) convert the norm \(\|\vx^{P^\perp}\|\) to a virtual logit by rescaling;
and (3) output the softmax probability of the virtual logit as the ViM score.
Below we give more details.
Recall the notations: \(C\) is the number of classes, \(N\) is the feature dimension, and \(\mW\) and \(\vb\) are the classification weight and bias, respectively.

\paragraph{Principal Subspace and Residual}
Firstly we offset the feature space by a vector \(\vo=-(\mW^T)^+\vb\) so that it is bias-free in the computation of logits as \cref{eq:bias-free}.
The principal subspace \(P\) is defined by the training set \(\mX\), where rows are features in the new coordinate system with origin \(\vo\).
Suppose the eigendecomposition on the matrix \(\mX^T \mX\) is
\begin{equation}\label{eq:eigen}
    \mX^T \mX = \mQ \mLambda \mQ^{-1},
\end{equation}
where eigenvalues in \(\mLambda\) are sorted decreasingly,
then the span of the first \(D\) columns is the \(D\)-dimensional principal subspace \(P\).
The residual \(\vx^{P^\perp}\) is the projection of \(\vx\) onto \(P^\perp\),
Let the \((D+1)\)-th column to the last column of \(\mQ\) in \cref{eq:eigen} be a new matrix \(\mR\in\R^{N\times(N-D)}\),
then \(\vx^{P^\perp}=\mR\mR^T\vx\).
The residual \(\vx^{P^\perp}\) is sent to the next step.

\paragraph{Virtual-logit Matching}
The virtual logit
\begin{equation}\label{eq:vlogit}
    l_0:=\alpha\|\vx^{P^\perp}\|=\alpha\sqrt{\vx^T\mR\mR^T\vx}
\end{equation}
is the norm of the residual rescaled by a per-model constant \(\alpha\).
The norm \(\|\vx^{P^\perp}\|\) cannot be used as a new logit directly since the latter softmax will normalize over the exponential of logits and thus is very sensitive to the scale of logits.
If the residual is very small compared to the largest logit, then after the softmax the residual will be buried in the noise of logits.
To match the scales of the virtual logit, we compute the average norm of the virtual logit on the training set and also the mean of the maximum logit on the training set, then
\begin{equation}\label{eq:alpha}
    \alpha := \frac{\sum_{i=1}^K \max_{j=1,\dots,C} \{l^i_j\}}{\sum_{i=1}^K \|\vx_i^{P^\perp}\|},
\end{equation}
where \(\vx_1, \vx_2, \ldots , \vx_K\) are uniformly sampled \(K\) training examples,
and \(l^i_j\) is the \(j\)-th logit of \(\vx_i\).
In this way, on average, the scale of the virtual logit is the same as the maximum of the original logits.

\paragraph{The ViM Score}
We append the virtual logit to the original logits and compute the softmax.
The probability corresponding to the virtual logit is defined as ViM\@.
Mathematically,
let the \(i\)-th logit of \(\vx\) be \(l_i\), and
then the score is
\begin{equation}
    \text{ViM}(\vx) = \frac{e^{\alpha\sqrt{\vx^T\mR\mR^T\vx}}}{\sum_{i=1}^C e^{l_i} + e^{\alpha\sqrt{\vx^T\mR\mR^T\vx}}}.
\end{equation}
This equation reveals that two factors affect the ViM score:
if its original logits are larger, then it is less of an OOD example;
while if the norm of residual is larger, it is more likely to be OOD\@.
The computational overhead is comparable to the last fully-connected layer (mapping from feature to logit) in the classification network, which is small.

\paragraph{Connection to Existing Methods}

Note that applying a strictly increasing function to the scores does not affect the OOD evaluation.
Apply the function \(t(x) = -\ln\left(\frac{1}{x}-1\right)\) to the ViM score,
then we have an equivalent expression
\begin{equation}\label{eq:connection}
    \alpha\|\vx^{P^\perp}\| - \ln\sum_{i=1}^C e^{l_i}.
\end{equation}
The first term is the virtual logit in \cref{eq:vlogit} while the second term is the energy score~\cite{energyood20nips}.
ViM completes the energy method by feeding extra residual information from features.
The performance is much superior to energy and residual.

\section{OpenImage-O Dataset}\label{sec:openimage}

We build a new OOD dataset called OpenImage-O for the ID dataset ImageNet-1K.
It is manually annotated, comes with a naturally diverse distribution, and has a large scale with 17,632 images.
It is built to overcome several shortcomings of existing OOD benchmarks.
OpenImage-O is selected image-by-image from the test set of OpenImage-V3, including 125,436 images collected from Flickr without a predefined list of class names or tags, leading to natural class statistics and avoiding an initial design bias.

\paragraph{Necessity for Image-Level Annotation}
Some previous works on large-scale OOD detection select a portion of other datasets solely based on class labels.
While class-level annotation costs less, the resulting dataset might be much noisier than expected.
For example, the Places and the SUN dataset selected by~\cite{huang2021mos} have a large portion of images that are indistinguishable from ID samples.
Another example is the Texture~\cite{cimpoi14describing,huang2021mos}, in which the \emph{bubbly} texture overlaps with the \emph{bubble} class in ImageNet.
Thus creating OOD datasets by querying tags is not reliable and per-image human inspection is needed for the confirmation of validity.

\paragraph{Hackability of Small Coverage}
If the OOD dataset has a central topic such as the Texture, featuring a less diverse distribution, then it might be easy to be ``hacked''.
In Tab.~2, the gap between the highest and the average AUROC over nine methods for BiT are: OpenImage-O 5.61, iNaturalist 6.06, Texture 10.52, and ImageNet-O 14.39.
Having larger gaps implies that the dataset is easier to improve.

\paragraph{Construction Process of OpenImage-O}
We construct the OpenImage-O based on the OpenImage-v3 dataset~\cite{openimages}.
For every image in its testing set,
we let human labelers to determine whether it is an OOD sample.
To assist labeling, we simplified the task as distinguishing the image from the top-10 categories predicted by an ImageNet-1K classification model, i.e., the image is OOD if it does not belong to any of the 10 categories.
Category labels as well as the most similar image to the test image in each category, measured by cosine similarity in the feature space, were presented for visualization.
To further improve the annotation quality, we design several schemes:
(1) Labelers can choose ``Difficult'', if they cannot decide whether the image belongs to any of the 10 categories;
(2) Each image was labeled by at least two labelers independently, and we took the set of OOD images having consensus from the two;
(3) Random inspection was performed to guarantee the quality.

\section{Experiment}\label{sec:exp}

In this section, we compare our algorithm with state-of-the-art OOD detection algorithms.
Following the prior work on large-scale OOD detection, we choose ImageNet-1K as the ID dataset.
We benchmark the algorithms using both the CNN-based and the transformer-based models.
Detailed experimental settings are as follows.

\paragraph{OOD Datasets}

Four OOD datasets (\cref{tab:datasets}) are used to comprehensively benchmark the algorithms.
OpenImage-O is our newly collected large-scale OOD dataset.
Texture~\cite{cimpoi14describing} consists of natural textural images and we removed four categories (\emph{bubbly, honeycombed, cobwebbed, spiralled}) that overlapped with ImageNet.
iNaturalist~\cite{van2018inaturalist} is a fine-grained species classification dataset.
We use the subset from~\cite{huang2021mos}.
Images in ImageNet-O~\cite{hendrycks2021natural} are adversarially filtered so that they can fool OOD detectors.

\paragraph{Evaluation Metrics}
Two commonly used metrics are reported.
The AUROC is a threshold-free metric that computes the area under the receiver operating characteristic curve.
Higher value indicates better detection performance.
FPR95 is short for FPR@TPR95, which is the false positive rate when the true positive rate is 95\%.
The smaller FPR95 the better.
We report both their numbers in percentage.

\paragraph{Experiment Settings}
BiT (Big Transfer)~\cite{kolesnikov2020big} is a variant of ResNet-v2, which employs group normalization and weight standardization.
The BiT-S model series is pre-trained on ImageNet-1K, and we take the officially released checkpoint of BiT-S-R101\(\times\)1 for experiments.
ViT (Vision Transformer)~\cite{dosovitskiy2021an} is a transformer-based image classification model which treats images as sequences of patches.
We use the officially released ViT-B/16 model, which is pre-trained on ImageNet-21K and fine-tuned on ImageNet-1K.
Since the compared algorithms do not require re-training, the ID accuracies are not affected.
Results on more model architectures, including CNN-based RepVGG~\cite{ding2021repvgg}, ResNet-50d~\cite{he2019bag}, and transformer based
Swin~\cite{liu2021swin} and DeiT~\cite{pmlr111}, are listed in \cref{sec:more}.
Their pre-trained weights are obtained from the timm repo~\cite{rw2019timm}.
When estimating the principal space, \(K=200,000\) images are randomly sampled from the training set.
For features spaces with dimension \(N>1500\), we set the dimension of principal space to \(D=1000\), and set \(D=512\) otherwise.

\begin{table*}
    \centering
    \setlength\tabcolsep{2.3 pt}
    \begin{tabular}{@{}l@{}llc@{}lc@{}lc@{}lc@{}lc@{}lc@{}lc@{}lc@{}}
        \toprule
        \multirow{2}{*}{\begin{tabular}[c]{@{}c@{}}\textbf{Model}\end{tabular}}
         & \multirow{2}{*}{\begin{tabular}[c]{@{}c@{}}\textbf{~~~Method}\end{tabular}} & \multirow{2}{*}{\begin{tabular}[c]{@{}c@{}}\textbf{Source}\end{tabular}} & \multicolumn{2}{c}{\textbf{OpenImage-O}} & \multicolumn{2}{c}{\textbf{Texture}} & \multicolumn{2}{c}{\textbf{iNaturalist}} & \multicolumn{2}{c}{\textbf{ImageNet-O}} & \multicolumn{2}{c}{\textbf{Average}}                                                                                                                                                        \\
         &                                            &                                            & {\small AUROC\(\uparrow\)}               & {\small FPR95\(\downarrow\)}         & {\small AUROC\(\uparrow\)}               & {\small FPR95\(\downarrow\)}            & {\small AUROC\(\uparrow\)}           & {\small FPR95\(\downarrow\)} & {\small AUROC\(\uparrow\)} & {\small FPR95\(\downarrow\)} & {\small AUROC\(\uparrow\)} & {\small FPR95\(\downarrow\)} \\
        \cmidrule(r){1-3}\cmidrule(lr){4-11}\cmidrule(l){12-13}
        \multirow{9}{*}{\begin{tabular}[c]{@{}c@{}}BiT\end{tabular}}
         & {MSP}~\cite{msp17iclr}                     & prob                                       & \(84.16\)                                & \(73.72\)                            & \(79.80\)                                & \(76.65\)                               & {\(87.92\)}                          & \(64.09\)                    & \(57.12\)                  & \(96.85\)                    & \(77.25\)                  & \(77.83\)                    \\
         & {Energy}~\cite{energyood20nips}            & logit                                      & \(84.77\)                                & \(73.42\)                            & \(81.09\)                                & \(73.91\)                               & \(84.47\)                            & \(74.98\)                    & \(63.59\)                  & \(96.40\)                    & \(78.48\)                  & \(79.68\)                    \\
         & {ODIN}~\cite{odin18iclr}                   & prob+grad                                  & \(85.64\)                                & \(72.83\)                            & \(81.60\)                                & \(74.07\)                               & \(86.73\)                            & \(70.75\)                    & \(63.00\)                  & \(96.85\)                    & \(79.24\)                  & \(78.63\)                    \\
         & {MaxLogit}~\cite{hendrycks2019scaling}     & logit                                      & {\(85.67\)}                              & \(72.68\)                            & \(81.66\)                                & \(73.72\)                               & \(86.76\)                            & \(70.59\)                    & \(63.01\)                  & \(96.85\)                    & \(79.27\)                  & \(78.46\)                    \\
         & {KL Matching}~\cite{hendrycks2019scaling}  & prob                                       & \underline{\(88.96\)}                    & \(51.51\)                            & \(86.92\)                                & \(51.05\)                               & \boldmath{\(92.95\)}                 & \boldmath{\(33.28\)}         & \(65.68\)                  & \(86.65\)                    & \(83.63\)                  & \(55.62\)                    \\
         & {Residual\(^\dagger\)}                     & feat                                       & \(80.58\)                                & \(67.85\)                            & \underline{\(97.66\)}                    & \(11.16\)                               & \(76.76\)                            & \(80.41\)                    & \underline{\(81.57\)}      & \(65.50\)                    & \(84.14\)                  & \(56.23\)                    \\
         & {ReAct}~\cite{sun2021tone}                 & feat+logit                                 & \underline{\(88.94\)}                    & \(54.97\)                            & {\(90.64\)}                              & \(50.25\)                               & \underline{\(91.45\)}                & \(48.60\)                    & {\(67.07\)}                & \(91.70\)                    & \underline{\(84.53\)}      & \(61.38\)                    \\
         & {Mahalanobis}~\cite{mahananobis18nips}     & feat+label                                 & \(83.10\)                                & \(64.32\)                            & \underline{\(97.33\)}                    & \(14.05\)                               & \(85.70\)                            & \(64.95\)                    & \underline{\(80.37\)}      & \(70.05\)                    & \underline{\(86.62\)}      & \(53.34\)                    \\
         & \textbf{ViM (Ours)}                        & feat+logit                                 & \boldmath{\(91.54\)}                     & \boldmath{\(43.96\)}                 & \boldmath{\(98.92\)}                     & \boldmath{\(~~4.69\)}                   & \underline{\(89.30\)}                & \(55.71\)                    & \boldmath{\(83.87\)}       & \boldmath{\(61.50\)}         & \boldmath{\(90.91\)}       & \boldmath{\(41.46\)}         \\
        \cmidrule(r){1-3}\cmidrule(lr){4-11}\cmidrule(l){12-13}
        \multirow{9}{*}{\begin{tabular}[c]{@{}c@{}}ViT\end{tabular}}
         & {MSP}~\cite{msp17iclr}                     & prob                                       & \(92.53\)                                & \(34.18\)                            & \(87.10\)                                & \(48.55\)                               & \(96.11\)                            & \(19.04\)                    & \(81.86\)                  & \(64.85\)                    & \(89.40\)                  & \(41.65\)                    \\
         & {Energy}~\cite{energyood20nips}            & logit                                      & {\(97.11\)}                              & \(14.04\)                            & \underline{\(93.39\)}                    & \(28.22\)                               & {\(98.66\)}                          & \(~~6.16\)                   & {\(90.46\)}                & \(41.30\)                    & \(94.90\)                  & \(22.43\)                    \\
         & {ODIN}~\cite{odin18iclr}                   & prob+grad                                  & \(96.86\)                                & \(15.68\)                            & \(93.01\)                                & \(30.60\)                               & \(98.57\)                            & \(~~6.58\)                   & \(89.85\)                  & \(44.15\)                    & \(94.57\)                  & \(24.25\)                    \\
         & {MaxLogit}~\cite{hendrycks2019scaling}     & logit                                      & \(96.87\)                                & \(15.68\)                            & \(93.01\)                                & \(30.60\)                               & \(98.57\)                            & \(~~6.58\)                   & \(89.85\)                  & \(44.15\)                    & \(94.57\)                  & \(24.25\)                    \\
         & {KL Matching}~\cite{hendrycks2019scaling}  & prob                                       & \(93.80\)                                & \(28.49\)                            & \(88.76\)                                & \(44.09\)                               & \(96.88\)                            & \(14.79\)                    & \(84.12\)                  & \(55.70\)                    & \(90.89\)                  & \(35.77\)                    \\
         & {Residual\(^\dagger\)}                     & feat                                       & \(92.72\)                                & \(32.63\)                            & \(92.21\)                                & \(33.80\)                               & \(98.57\)                            & \(~~6.63\)                   & \(88.23\)                  & \(47.85\)                    & \(92.93\)                  & \(30.23\)                    \\
         & {ReAct}~\cite{sun2021tone}                 & feat+logit                                 & \underline{\(97.38\)}                    & \(13.50\)                            & {\(93.34\)}                              & \(28.49\)                               & \underline{\(99.00\)}                & \(~~4.31\)                   & \underline{\(90.71\)}      & \(42.60\)                    & \underline{\(95.11\)}      & \(22.22\)                    \\
         & {Mahalanobis}~\cite{mahananobis18nips}     & feat+label                                 & \underline{\(97.48\)}                    & \(13.54\)                            & \underline{\(94.24\)}                    & \(25.17\)                               & \boldmath{\(99.54\)}                 & \boldmath{\(~~2.12\)}        & \boldmath{\(92.81\)}       & \(36.95\)                    & \underline{\(96.02\)}      & \(19.45\)                    \\
         & \textbf{ViM (Ours)}                        & feat+logit                                 & \boldmath{\(97.61\)}                     & \boldmath{\(12.61\)}                 & \boldmath{\(95.34\)}                     & \boldmath{\(20.31\)}                    & \underline{\(99.41\)}                & \(~~2.60\)                   & \underline{\(92.55\)}      & \boldmath{\(36.75\)}         & \boldmath{\(96.23\)}       & \boldmath{\(18.07\)}         \\
        \bottomrule
    \end{tabular}
    \caption{
        OOD detection for ViM and baseline methods.
        The ID dataset is ImageNet-1K, and OOD datasets are OpenImage-O, Texture, iNaturalist and ImageNet-O.
        Both metrics AUROC and FPR95 are in percentage.
        A pre-trained BiT-S-R101\(\times\)1 model and a pre-trained ViT-B/16 model is tested.
        The best method is emphasized in bold, and the 2nd and 3rd ones are underlined.
        ODIN needs backpropagation for producing input perturbations, so it is \emph{prob+grad}.
        ReAct clips feature and uses Energy subsequently, so it is \emph{feat+logit}.
        Mahalanobis need gt labels to compute the class-wise mean feature, so it is \emph{feat+label}.
        \(^\dagger\): Residual is defined in \cref{eq:residual}.
    }\label{tab:grand}
\end{table*}

\paragraph{Baseline Methods}
We compare ViM with eight baselines that do not require fine-tuning.
They are {MSP}~\cite{msp17iclr}, {Energy}~\cite{energyood20nips}, {ODIN}~\cite{odin18iclr}, {MaxLogit}~\cite{hendrycks2019scaling}, {KL Matching}~\cite{hendrycks2019scaling}, {Residual}, {ReAct}~\cite{sun2021tone} and {Mahalanobis}~\cite{mahananobis18nips}.
For Mahalanobis,
we followed the setting in~\cite{nearood21arxiv}, which uses only the final feature instead of an ensemble of multiple layers~\cite{mahananobis18nips,huang2021mos}.
For ReAct, we use the Energy+ReAct setting with rectification percentile \(p=99\).
The Residual is defined in \cref{eq:residual}.

\subsection{Results on BiT}\label{sec:bit}

We present the results of the BiT model at the first half of \cref{tab:grand}.
The best AUROC is shown in bold and the second and third place ones are shown with underlines.

\paragraph{ViM \versus Baselines}
On three datasets, including OpenImage-O, Texture, and ImageNet-O, ViM achieves the largest AUROC and the smallest FPR95.
On average ViM has 90.91\% AUROC, which surpasses the second place by 4.29\%.
The average FPR95 is also the lowest among them.
In particular, regarding \cref{eq:connection}, an interpretation of ViM in terms of the Residual score and the Energy score, the results show that ViM is significantly better than the two methods on all datasets.
This indicates that ViM non-trivially combined the OOD information in Residual and in Energy.
However, on iNaturalist, ViM is only on the third place.
We hypothesize that its moderate performance on iNaturalist relates to how much information is contained in the residual, because
iNaturalist has the smallest average residual norm among four OOD datasets (iNaturalist 4.65, OpenImage-O 5.04, ImageNet-O 5.16, and Texture 8.16).

\paragraph{Effect of Information Source}
For OOD detection performances on BiT model, \cref{tab:grand} shows an interesting pattern regarding the information source.
If feature variations in the null space are absent, such as in methods that rely on logits and softmax, performances on Texture and ImageNet-O are restricted.
For example, on the Texture dataset, the best performing method that relies on logit and softmax is KL Matching, which has 86.92\% AUROC and is far behind ViM, Mahalanobis, and Residual, which operate on the feature space.
In contrast, if the class-dependent information is dropped, such as in the Residual method, performances in iNaturalist and OpenImage-O are also limited.
The proposed ViM score, however, is competent regardless of dataset types.

\subsection{Results on ViT}

\cite{nearood21arxiv} has discussed the benefit of large-scale pre-trained transformers on OOD tasks.
However, their experiments are conducted on CIFAR100/10 and only two baseline methods are compared.
We provide a comprehensive OOD evaluation on ImageNet-1K over a wide range of methods in the second half of \cref{tab:grand}.

\paragraph{ViM \versus Baselines}
The two best-performing methods for the ViT model are ViM and Mahalanobis.
Their AUROCs are close on all four datasets.
However, Mahalanobis needs to compute the class-wise Mahalanobis distance, which makes its computation costly.
In contrast, our method is lightweight and fast.
Four methods, ReAct, Energy, MaxLogit, and ODIN, are the second best ones,
and the remaining three methods have relatively low AUROCs.

\begin{table}
    \centering
    \footnotesize
    \setlength\tabcolsep{1.2 pt}
    \begin{tabular}{lcccccccc}
        \toprule
        \multirow{2}{*}{\begin{tabular}[c]{@{}c@{}}\textbf{Method}\end{tabular}} & \multicolumn{2}{c}{\textbf{RepVGG}~\cite{ding2021repvgg}} & \multicolumn{2}{c}{\textbf{Res50d}~\cite{he2019bag}} & \multicolumn{2}{c}{\textbf{Swin}~\cite{liu2021swin}} & \multicolumn{2}{c}{\textbf{DeiT}~\cite{pmlr111}}                                                                                                         \\
                                                   & {\small A\(\uparrow\)}                                    & {\small F\(\downarrow\)}                             & {\small A\(\uparrow\)}                               & {\small F\(\downarrow\)}                         & {\small A\(\uparrow\)} & {\small F\(\downarrow\)} & {\small A\(\uparrow\)} & {\small F\(\downarrow\)} \\
        \midrule
        MSP                                        & \(78.10\)                                                 & \(70.55\)                                            & \(77.99\)                                            & \(67.96\)                                        & {\(87.57\)}            & {\(43.44\)}              & {\(79.48\)}            & {\(66.43\)}              \\
        Energy                                     & \(76.38\)                                                 & \(78.99\)                                            & \(71.08\)                                            & \(78.39\)                                        & {\(87.77\)}            & {\(35.08\)}              & {\(72.80\)}            & {\(70.14\)}              \\
        ODIN                                       & \(77.72\)                                                 & \(72.68\)                                            & \(75.27\)                                            & \(68.56\)                                        & {\(88.00\)}            & {\(36.58\)}              & {\(77.13\)}            & \boldmath{\(63.92\)}     \\
        MaxLogit                                   & \(77.56\)                                                 & \(73.50\)                                            & \(75.39\)                                            & \(69.34\)                                        & {\(88.40\)}            & {\(35.28\)}              & {\(76.79\)}            & {\(64.49\)}              \\
        \scalebox{.9}[1.0]{KL Matching}            & \(81.35\)                                                 & \(61.65\)                                            & \(82.72\)                                            & \(64.41\)                                        & {\(88.87\)}            & {\(46.99\)}              & {\(83.49\)}            & {\(64.80\)}              \\
        Residual                                   & \underline{\(84.19\)}                                     & \(59.00\)                                            & \underline{\(87.01\)}                                & \(58.55\)                                        & \underline{\(92.88\)}  & {\(37.38\)}              & \underline{\(84.15\)}  & {\(74.13\)}              \\
        ReAct                                      & {\(49.14\)}                                               & \(98.96\)                                            & {\(82.93\)}                                          & \(58.63\)                                        & \(90.17\)              & \(31.36\)                & \(77.37\)              & \(67.00\)                \\
        \scalebox{.95}[1.0]{Mahalanobis}           & \underline{\(86.07\)}                                     & \(59.39\)                                            & \underline{\(88.33\)}                                & \(55.70\)                                        & \underline{\(92.16\)}  & {\(40.39\)}              & \underline{\(85.03\)}  & {\(73.18\)}              \\
        ViM (Ours)                                 & \boldmath{\(87.81\)}                                      & \boldmath{\(50.50\)}                                 & \boldmath{\(89.22\)}                                 & \boldmath{\(52.61\)}                             & \boldmath{\(94.11\)}   & \boldmath{\(31.04\)}     & \boldmath{\(85.25\)}   & {\(69.95\)}              \\
        \bottomrule
    \end{tabular}
    \caption{
        Results on RepVGG, ResNet50-d, Swin and DeiT.
        Due to space limitation, only their average AUROC (A\(\uparrow\)) and average FPR95 (F\(\downarrow\)) are reported.
        The numbers are in percentage.
        All models are using pre-trained weights taken from timm~\cite{rw2019timm}.
    }\label{tab:other}
\end{table}

\paragraph{Difference between ViT and BiT}
Since the ViT model is pre-trained on the ImageNet-21K dataset, the semantics it has seen is much larger than the BiT model.
The OOD performance is relatively saturated.
Although on most OOD datasets ViT is significantly better than BiT, we observe that ViT performs less competitively on the Texture dataset.
We hypothesize that it is related to the observation in~\cite{raghu2021vision} that higher layers of ViT maintain spatial location information more faithfully than ResNets.
ViT has high responses for local patches.
However, textural images with similar local patches but not revealing the whole object are regarded as OOD of ImageNet (see example images in~\cref{fig:hack_dataset}).

\subsection{Results on More Model Architectures}\label{sec:more}

We show more results on a variety of model architectures.
In particular, we choose two CNN-based models RepVGG~\cite{ding2021repvgg} and ResNet-50d~\cite{he2019bag} and two transformer-based
models Swin Transformer~\cite{liu2021swin} and DeiT~\cite{pmlr111}.
Their average AUROCs and average FPR95s over the four OOD datasets are listed in \cref{tab:other}.
It is shown that ViM is robust to model architecture changes.
The detailed experiment setting and results are in the supplementary materials.

\subsection{The Effect of Hyperparameter}

\begin{figure}[t]
  \begin{center}
    \includegraphics[width=0.24\textwidth]{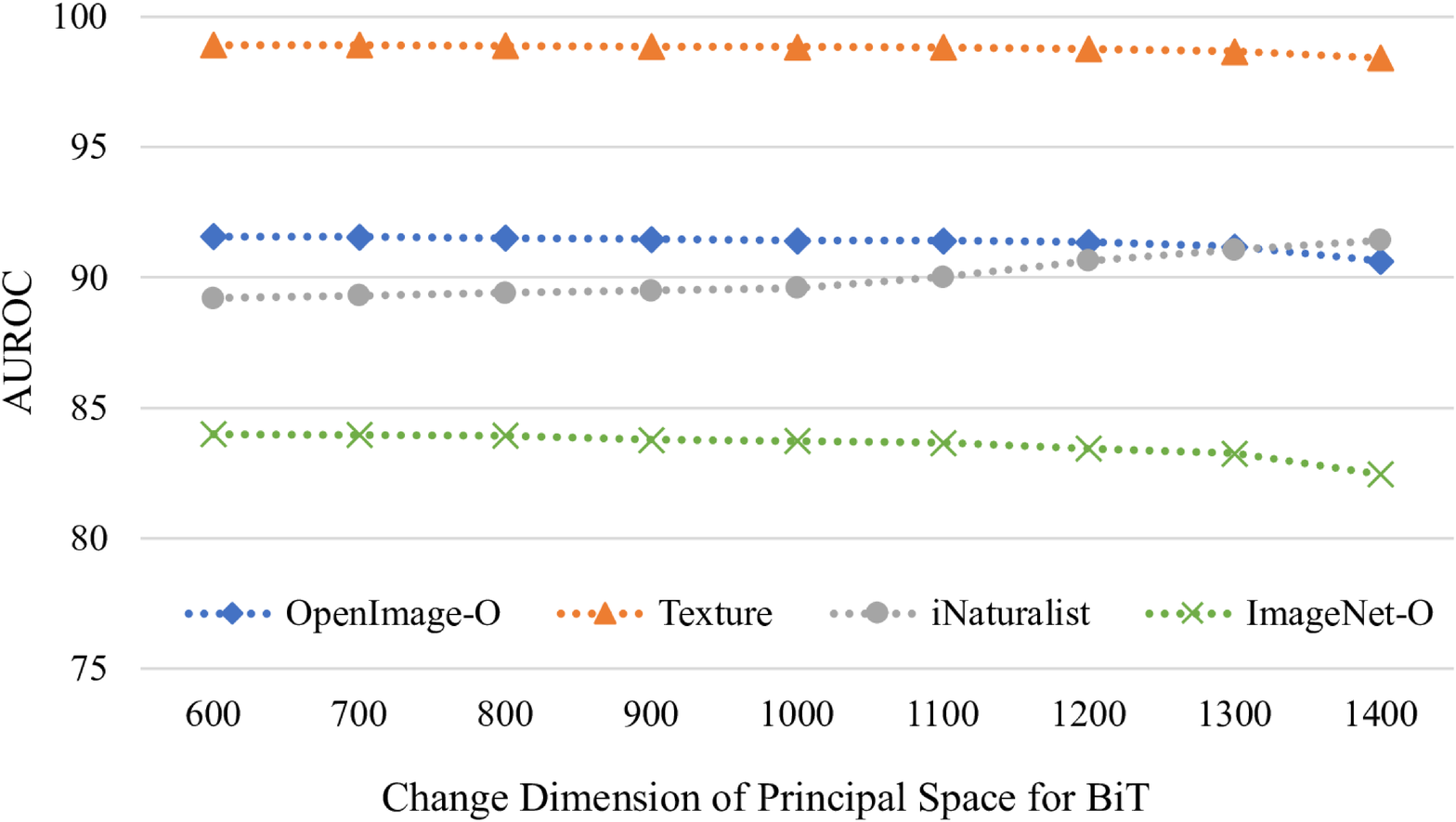}
    \includegraphics[width=0.23\textwidth]{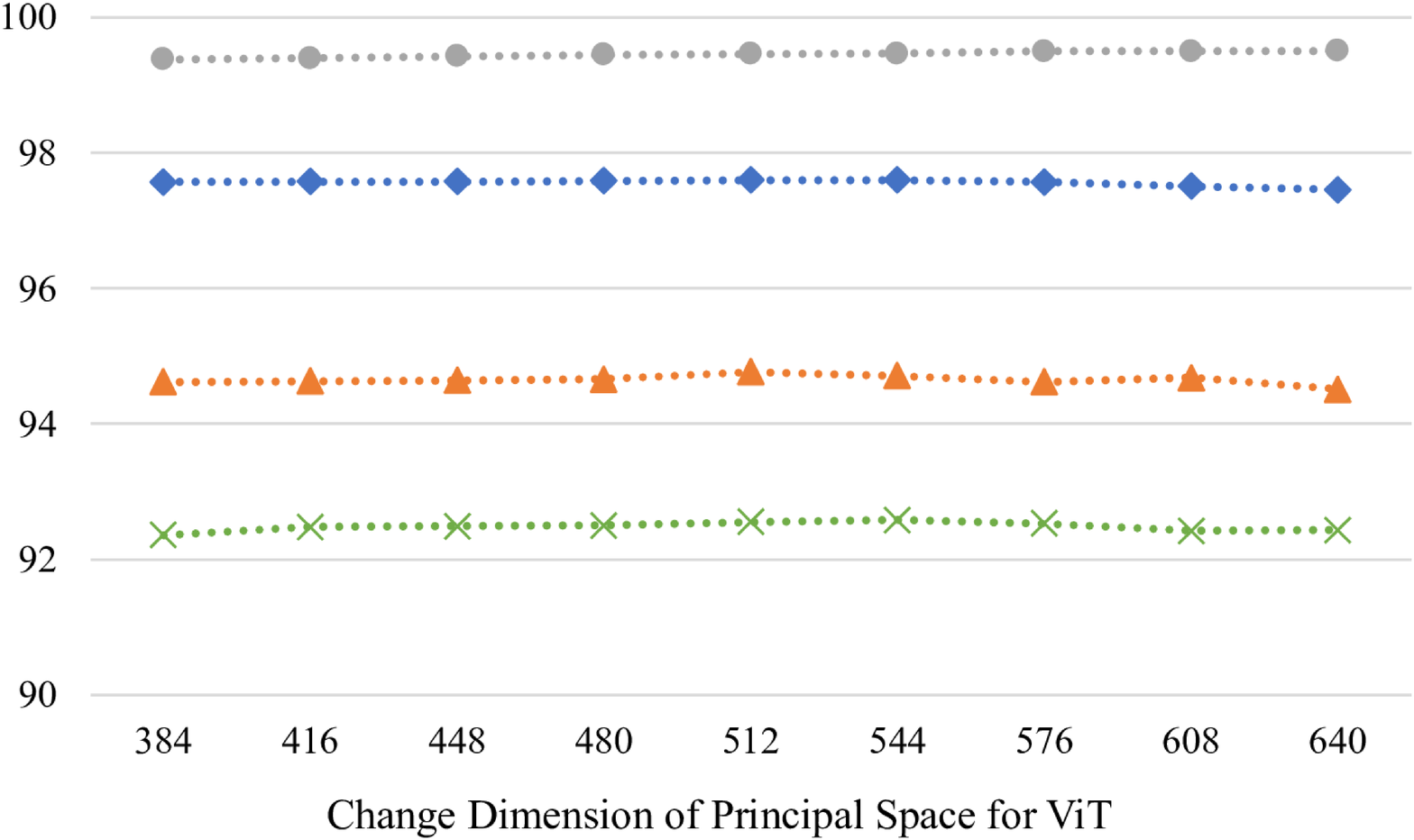}
  \end{center}
  \caption{
    Robustness against principal space dimension.
    Left is BiT and right is ViT.
    The performance changes are small when \(D\) varies in a wide range of values.
  }\label{fig:tune_d}
\end{figure}

\begin{figure}[t]
  \begin{center}
    \includegraphics[width=0.24\textwidth]{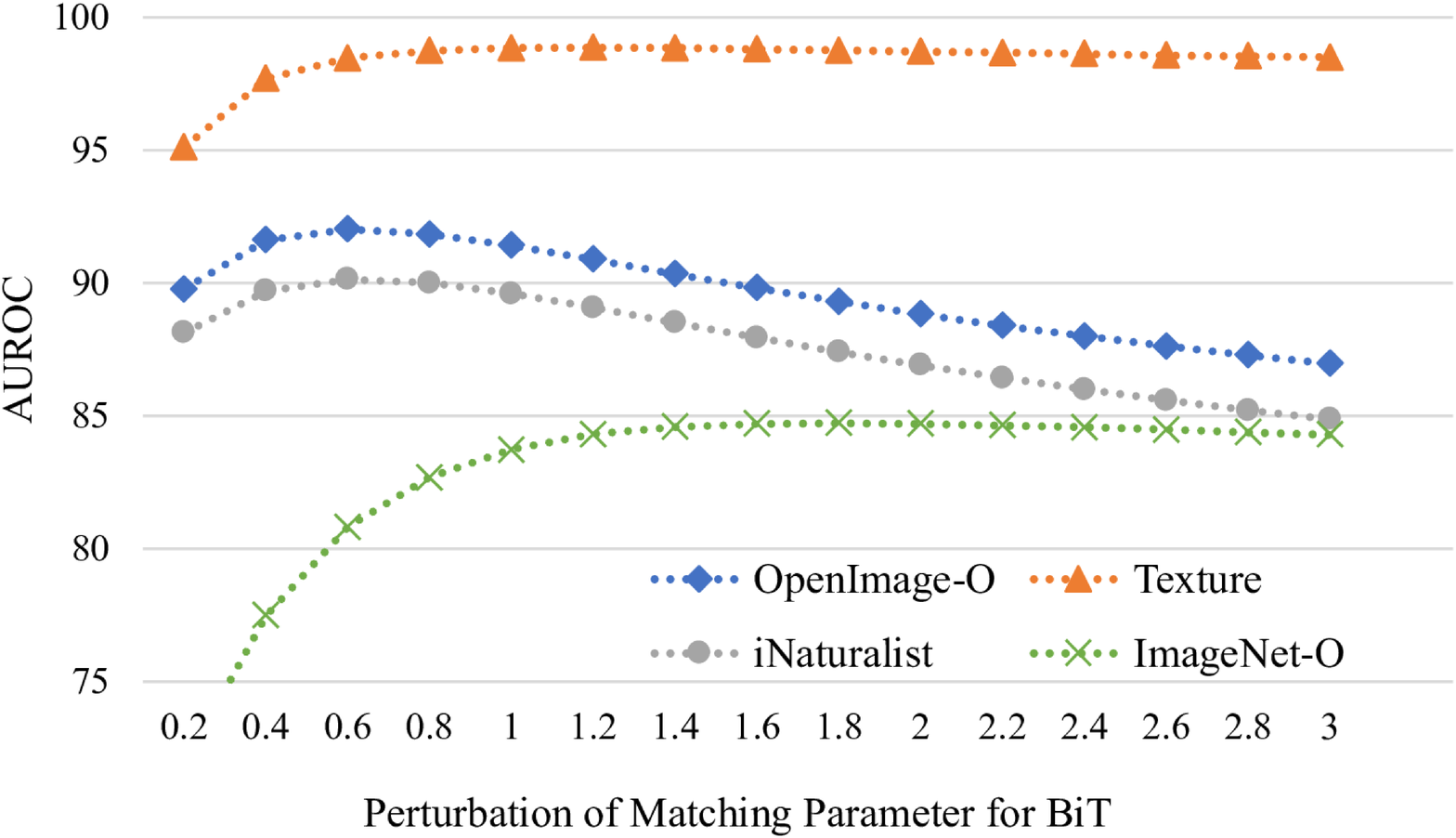}
    \includegraphics[width=0.23\textwidth]{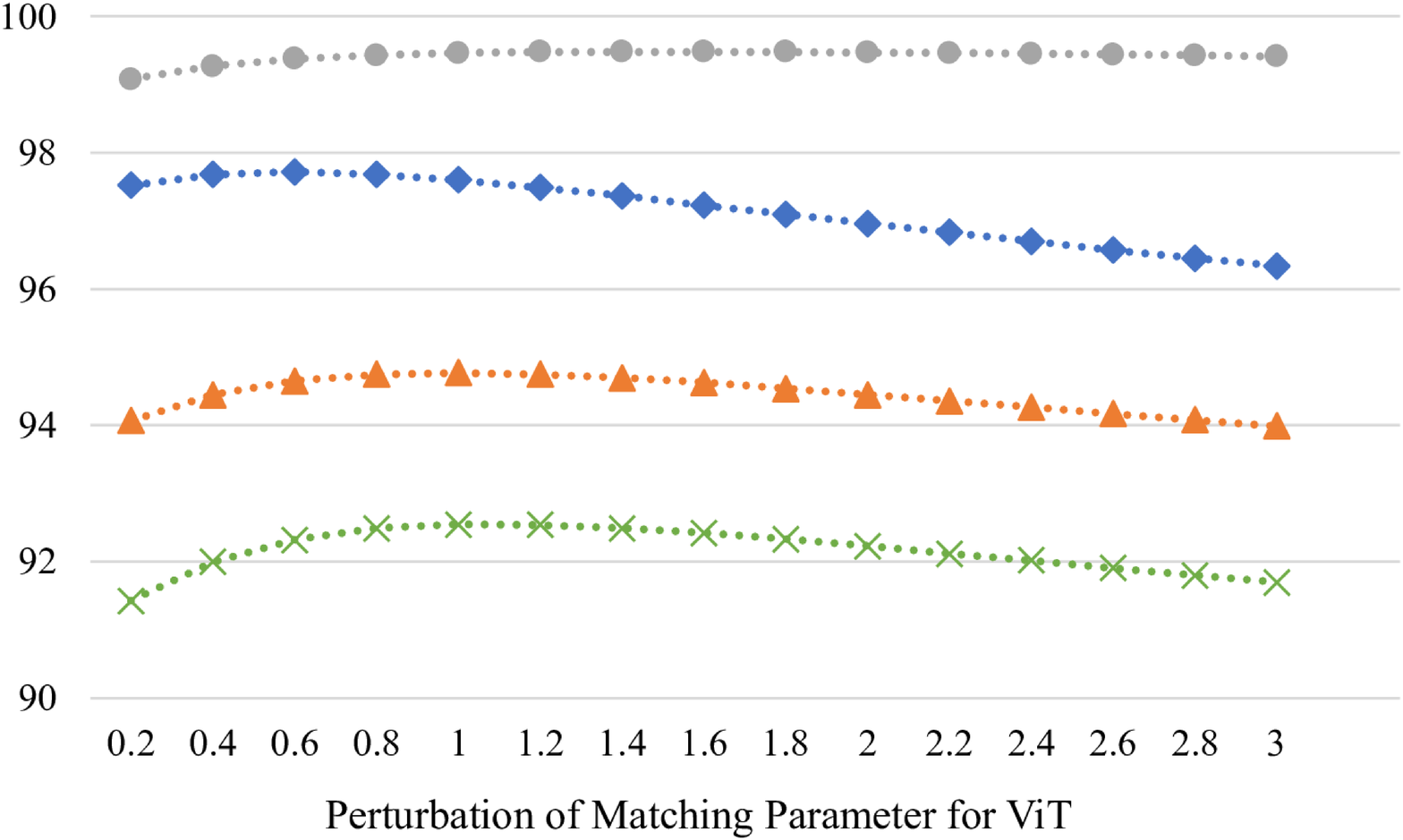}
  \end{center}
  \caption{
    Perturbation of \(\alpha\) by multiplying a factor.
    Left is BiT, and right is ViT.
    For both models, the proposed matching parameter fits well for the trends.
  }\label{fig:tune_alpha}
\end{figure}

\paragraph{The Dimension \(D\) of Principal Space}
In~\cite{onedim21cvpr} the feature of each class is represented by a 1-dimensional subspace, so a natural choice for the dimension \(D\) of principal space is the number of classes \(C\).
For models like ViT whose feature dimension \(N\) may be less than the number of classes \(C\), we empirically suggest taking a number in the range \([N/3, 2N/3]\).
We show in \cref{fig:tune_d} that our method is robust to the selection of dimensions.
However, if the application permits, one can adjust this parameter according to a hold-out OOD dataset.
In our experiments, we set \(D = 1000\) for BiT and \(D=512\) for ViT.

\paragraph{The Matching Parameter \(\alpha\)}
The matching parameter controls the relative importance of the trade-off between different OOD features.
Since OOD distribution is unknown, we suggest keeping them to be of equal importance.
This is how \(\alpha\) is defined in \cref{eq:alpha}.
It is easy to tune the parameter to fit some types of OOD datasets, but it is hard to improve all datasets at the same time.
We show the result of perturbing the matching parameter by multiplying a factor in \cref{fig:tune_alpha}.
If the multiple is larger, then information from the feature space is given more weight.
Otherwise, information from logits is given more importance.
Overall the best choice is no perturbation, suggesting that the defined \(\alpha\) is a good choice.

\subsection{The Effect of Grouping}\label{sec:cluster}

\begin{table}
    \centering
    \footnotesize
    \setlength\tabcolsep{1.3pt}
    \begin{tabular}{lccccccccc}
        \toprule
        \multirow{2}{*}{\begin{tabular}[c]{@{}c@{}}\textbf{Method}\end{tabular}} & \multicolumn{2}{c}{\scalebox{.9}[1.0]{\textbf{OpenImage-O}}} & \multicolumn{2}{c}{\textbf{Texture}} & \multicolumn{2}{c}{\scalebox{.95}[1.0]{\textbf{iNaturalist}}} & \multicolumn{2}{c}{\scalebox{.9}[1.0]{\textbf{ImageNet-O}}}                                                                                                         \\
                                                   & {\small A\(\uparrow\)}                                       & {\small F\(\downarrow\)}             & {\small A\(\uparrow\)}                                        & {\small F\(\downarrow\)}                                    & {\small A\(\uparrow\)} & {\small F\(\downarrow\)} & {\small A\(\uparrow\)} & {\small F\(\downarrow\)} \\
        \midrule
        {MOS*}~\cite{huang2021mos}                 & \(89.14\)                                                    & \boldmath{\(41.97\)}                 & \(82.35\)                                                     & \(59.30\)                                                   & \boldmath{\(98.15\)}   & \boldmath{\(~~9.28\)}    & \(60.62\)              & \(86.65\)                \\
        \scalebox{.95}[1.0]{{MaxGroup} }           & \(84.75\)                                                    & \(71.22\)                            & \(80.42\)                                                     & \(77.87\)                                                   & \(89.50\)              & \(57.18\)                & \(63.93\)              & \(92.45\)                \\
        \scalebox{.95}[1.0]{{ViM+Group}}           & \boldmath{\(91.92\)}                                         & \(42.26\)                            & \boldmath{\(98.91\)}                                          & \boldmath{\(~~4.69\)}                                       & \(90.16\)              & \(52.74\)                & \boldmath{\(83.43\)}   & \boldmath{\(62.00\)}     \\
        \bottomrule
    \end{tabular}
    \caption{
        AUROC of methods with grouping information.
        A\(\uparrow\) is AUROC and F\(\downarrow\) is FPR95.
        All numbers are in percentage.
        BiT is used and the grouping is defined in~\cite{huang2021mos} based on taxonomy.
        * MOS needs fine-tuning while others do not.
    }\label{tab:cluster}
\end{table}

In addition, we also compare with MOS~\cite{huang2021mos}, which exploits grouping structure in large-scale semantic spaces.
Two methods are added to the comparison.
(1) \emph{MaxGroup} is the group version of MSP, which first obtains the group-wise probability by summing over the constituent classes, and then takes the maximum group probability as the ID score.
(2) \emph{ViM+Group} also takes the maximum group probability as the ID score, except that the probabilities are taken from the \((C+1)\) dimensional vector, with an extra ViM virtual class participating in the softmax normalization.

MaxGroup and ViM+Group are evaluated on the pre-trained weights of BiT, while MOS needs to fine-tune the model using group-based learning.
Results are shown in \cref{tab:cluster}.
We observe that
(1) the average AUROC of MaxGroup improves over the vanilla MSP from 77.25\% to 79.23\%, showing the usefulness of group information;
and (2)
both our original ViM and the group version of ViM are better than MOS on three of four datasets by large margins.

\subsection{Limitation of ViM}\label{sec:limitation}

As we have noticed in \cref{sec:bit}, ViM shows less performance gains on OOD datasets that have small residuals, such as iNaturalist.
Besides, the property that ViM does not need training is a double-edged sword.
It means that ViM is limited by the feature quality of the original network.

\section{Conclusion}\label{sec:conclusion}

In this paper, we present a novel OOD detection method: the Virtual-logit Matching (ViM) score.
It combines the information from both the feature space and the logits,
which provides the class-agnostic information and the class-dependent information, respectively.
Extensive experiments on the large-scale OOD benchmarks show the effectiveness and robustness of the method.
Especially, we tested ViM on both CNN-based models and transformer-based models, showing its robustness across model architectures.
To facilitate the evaluation of large-scale OOD detection, we create the OpenImage-O dataset for ImageNet-1K, which is of high-quality and large-scale.

\section*{Acknowledgement}
This work was supported in part by Innovation and Technology Commission of the Hong Kong Special Administrative Region, China (Enterprise Support Scheme under the Innovation and Technology Fund B/E030/18).
Haoqi Wang was also supported by the Technology Leaders of Tomorrow (TLT) Programme of HKSTP InnoAcademy.

% \clearpage
\appendix

\section{Detailed Information of Models (Sec. 6)}

In the experiment, we benchmarked a collection of deep classification models.
Their detailed information, including the specification, the architecture, the pre-train information, and the top-1 accuracy, is listed in \cref{tab:model_detail}.
To summarize, half of them are CNN-based, and half are transformer-based.
Vision Transformer and Swin Transformer are pre-trained on ImageNet-21K before training on the ImageNet-1K.

\begin{table*}[t]
    \centering
    \begin{tabular}{lllcc}
        \toprule
        \textbf{Model}               & \textbf{Specification}       & \textbf{Architecture} & \textbf{Pre-Trained Dataset} & \textbf{Top1 (\%)} \\
        \midrule
        BiT~\cite{kolesnikov2020big} & BiT-S-R101x1                 & CNN                   & ---                          & \(81.30 \)         \\
        ViT~\cite{dosovitskiy2021an} & ViT-B/16                     & Transformer           & ImageNet-21K                 & \(85.43\)          \\
        RepVGG~\cite{ding2021repvgg} & RepVGG-b3                    & CNN                   & ---                          & \(80.52\)          \\
        Res50d~\cite{he2019bag}      & ResNet-50d                   & CNN                   & ---                          & \(80.52\)          \\
        Swin~\cite{liu2021swin}      & Swin-base-patch4-window7-224 & Transformer           & ImageNet-21K                 & \(85.27\)          \\
        DeiT~\cite{pmlr111}          & DeiT-base-patch16-224        & Transformer           & ---                          & \(81.98\)          \\
        \bottomrule
    \end{tabular}
    \caption{
        Detailed information on the used models.
        The detailed specification and the top-1 accuracy of the model are provided.
        Three of them are CNN-based, and the other three are transformer-based.
        Both ViT and Swin Transformer are pre-trained on ImageNet-21K before training on ImageNet-1K, so their general OOD performances are much better than alternatives.
    }\label{tab:model_detail}
\end{table*}

\section{Detailed Results of Four Models (Sec. 6.3)}

\begin{table*}[t]
    \centering
    \setlength\tabcolsep{2.5pt}
    \renewcommand{\arraystretch}{1.2}
    \begin{tabular}{@{}l@{}llc@{}lc@{}lc@{}lc@{}lc@{}lc@{}lc@{}lc@{}}
        \toprule
        \multirow{2}{*}{\begin{tabular}[c]{@{}c@{}}\textbf{Model}\end{tabular}}
         & \multirow{2}{*}{\begin{tabular}[c]{@{}c@{}}\textbf{~~~Method}\end{tabular}} & \multirow{2}{*}{\begin{tabular}[c]{@{}c@{}}\textbf{Source}\end{tabular}} & \multicolumn{2}{c}{\textbf{OpenImage-O}}       & \multicolumn{2}{c}{\textbf{Texture}} & \multicolumn{2}{c}{\textbf{iNaturalist}}       & \multicolumn{2}{c}{\textbf{ImageNet-O}} & \multicolumn{2}{c}{\textbf{Average}}                                                                                                                                                                                                          \\
         &                                            &                                            & {\small \scalebox{.9}[1.0]{AUROC\(\uparrow\)}} & {\small FPR95\(\downarrow\)}         & {\small \scalebox{.9}[1.0]{AUROC\(\uparrow\)}} & {\small FPR95\(\downarrow\)}            & {\small \scalebox{.9}[1.0]{AUROC\(\uparrow\)}} & {\small FPR95\(\downarrow\)} & {\small \scalebox{.9}[1.0]{AUROC\(\uparrow\)}} & {\small FPR95\(\downarrow\)} & {\small \scalebox{.9}[1.0]{AUROC\(\uparrow\)}} & {\small FPR95\(\downarrow\)} \\
        \cmidrule(r){1-3}\cmidrule(lr){4-11}\cmidrule(l){12-13}
        \multirow{8}{*}{\begin{tabular}[c]{@{}c@{}}\scalebox{.9}[1.0]{RepVGG}~\cite{ding2021repvgg}\end{tabular}}
         & {MSP}~\cite{msp17iclr}                     & prob                                       & {\(85.06\)}                                    & {\(63.36\)}                          & {\(78.58\)}                                    & {\(72.62\)}                             & {\(87.11\)}                                    & {\(54.93\)}                  & {\(61.65\)}                                    & {\(91.30\)}                  & {\(78.10\)}                                    & {\(70.55\)}                  \\
         & {Energy}~\cite{energyood20nips}            & logit                                      & {\(83.64\)}                                    & {\(69.92\)}                          & {\(74.53\)}                                    & {\(82.97\)}                             & {\(83.92\)}                                    & {\(75.31\)}                  & {\(63.36\)}                                    & {\(87.75\)}                  & {\(76.36\)}                                    & {\(78.99\)}                  \\
         & {ODIN}~\cite{odin18iclr}                   & prob+grad                                  & {\(85.22\)}                                    & {\(63.48\)}                          & {\(76.77\)}                                    & {\(76.14\)}                             & {\(86.37\)}                                    & {\(61.40\)}                  & {\(62.50\)}                                    & {\(89.70\)}                  & {\(77.72\)}                                    & {\(72.68\)}                  \\
         & {MaxLogit}~\cite{hendrycks2019scaling}     & logit                                      & {\(84.81\)}                                    & {\(65.04\)}                          & {\(76.33\)}                                    & {\(76.86\)}                             & {\(86.22\)}                                    & {\(62.20\)}                  & {\(62.87\)}                                    & {\(89.90\)}                  & {\(77.56\)}                                    & {\(73.50\)}                  \\
         & {KL Matching}~\cite{hendrycks2019scaling}  & prob                                       & \underline{\(86.80\)}                          & {\(57.48\)}                          & {\(83.18\)}                                    & {\(62.09\)}                             & \underline{\(89.06\)}                          & \boldmath{\(42.07\)}         & {\(66.36\)}                                    & {\(84.95\)}                  & {\(81.35\)}                                    & {\(61.65\)}                  \\
         & {Residual\(^\dagger\)}                     & feat                                       & {\(82.51\)}                                    & {\(65.13\)}                          & \underline{\(93.05\)}                          & {\(28.66\)}                             & {\(86.09\)}                                    & {\(62.40\)}                  & \underline{\(75.11\)}                          & {\(79.80\)}                  & \underline{\(84.19\)}                          & {\(59.00\)}                  \\
         & {ReAct}~\cite{sun2021tone}                 & feat                                       & {\(46.08\)}                                    & \(99.65\)                            & {\(54.56\)}                                    & \(97.66\)                               & {\(47.18\)}                                    & \(99.88\)                    & {\(48.76\)}                                    & \(98.65\)                    & \(49.14\)                                      & \(98.96\)                    \\
         & {Mahalanobis}~\cite{mahananobis18nips}     & feat+label                                 & \underline{\(85.71\)}                          & {\(64.93\)}                          & \underline{\(92.71\)}                          & {\(32.03\)}                             & \underline{\(89.17\)}                          & {\(58.79\)}                  & \underline{\(76.68\)}                          & {\(81.80\)}                  & \underline{\(86.07\)}                          & {\(59.39\)}                  \\
         & \textbf{ViM (Ours)}                        & feat+logit                                 & \boldmath{\(89.27\)}                           & \boldmath{\(52.40\)}                 & \boldmath{\(93.69\)}                           & \boldmath{\(23.76\)}                    & \boldmath{\(91.35\)}                           & {\(46.79\)}                  & \boldmath{\(76.93\)}                           & \boldmath{\(79.05\)}         & \boldmath{\(87.81\)}                           & \boldmath{\(50.50\)}         \\
        \cmidrule(r){1-3}\cmidrule(lr){4-11}\cmidrule(l){12-13}
        \multirow{8}{*}{\begin{tabular}[c]{@{}c@{}}Res50d~\cite{he2019bag}\end{tabular}}
         & {MSP}~\cite{msp17iclr}                     & prob                                       & {\(84.50\)}                                    & {\(63.53\)}                          & {\(82.75\)}                                    & {\(64.40\)}                             & {\(88.58\)}                                    & {\(50.05\)}                  & {\(56.13\)}                                    & {\(93.85\)}                  & {\(77.99\)}                                    & {\(67.96\)}                  \\
         & {Energy}~\cite{energyood20nips}            & logit                                      & {\(75.95\)}                                    & {\(76.83\)}                          & {\(73.93\)}                                    & {\(75.31\)}                             & {\(80.50\)}                                    & {\(71.32\)}                  & {\(53.95\)}                                    & {\(90.10\)}                  & {\(71.08\)}                                    & {\(78.39\)}                  \\
         & {ODIN}~\cite{odin18iclr}                   & prob+grad                                  & {\(81.53\)}                                    & {\(64.49\)}                          & {\(80.21\)}                                    & {\(63.93\)}                             & {\(86.48\)}                                    & {\(52.58\)}                  & {\(52.87\)}                                    & {\(93.25\)}                  & {\(75.27\)}                                    & {\(68.56\)}                  \\
         & {MaxLogit}~\cite{hendrycks2019scaling}     & logit                                      & {\(81.50\)}                                    & {\(65.50\)}                          & {\(79.25\)}                                    & {\(66.20\)}                             & {\(86.42\)}                                    & {\(53.00\)}                  & {\(54.39\)}                                    & {\(92.65\)}                  & {\(75.39\)}                                    & {\(69.34\)}                  \\
         & {KL Matching}~\cite{hendrycks2019scaling}  & prob                                       & \underline{\(87.31\)}                          & {\(60.58\)}                          & {\(86.07\)}                                    & {\(61.36\)}                             & \boldmath{\(90.48\)}                           & \boldmath{\(47.22\)}         & {\(67.00\)}                                    & {\(88.50\)}                  & {\(82.72\)}                                    & {\(64.41\)}                  \\
         & {Residual\(^\dagger\)}                     & feat                                       & {\(87.64\)}                                    & {\(59.65\)}                          & \underline{\(94.62\)}                          & {\(25.89\)}                             & {\(84.63\)}                                    & {\(75.81\)}                  & \boldmath{\(81.15\)}                           & \boldmath{\(72.85\)}         & \underline{\(87.01\)}                          & {\(58.55\)}                  \\
         & {ReAct}~\cite{sun2021tone}                 & feat                                       & {\(85.30\)}                                    & \(60.79\)                            & {\(91.12\)}                                    & \(39.26\)                               & {\(87.27\)}                                    & \(56.03\)                    & {\(68.02\)}                                    & \(78.45\)                    & \(82.93\)                                      & \(58.63\)                    \\
         & {Mahalanobis}~\cite{mahananobis18nips}     & feat+label                                 & \underline{\(89.52\)}                          & {\(55.91\)}                          & \underline{\(94.15\)}                          & {\(28.22\)}                             & \underline{\(89.48\)}                          & {\(62.69\)}                  & \underline{\(80.15\)}                          & {\(76.00\)}                  & \underline{\(88.33\)}                          & {\(55.70\)}                  \\
         & \textbf{ViM (Ours)}                        & feat+logit                                 & \boldmath{\(90.76\)}                           & \boldmath{\(50.45\)}                 & \boldmath{\(95.84\)}                           & \boldmath{\(20.58\)}                    & \underline{\(89.26\)}                          & {\(64.59\)}                  & \underline{\(81.02\)}                          & {\(74.80\)}                  & \boldmath{\(89.22\)}                           & \boldmath{\(52.61\)}         \\
        \cmidrule(r){1-3}\cmidrule(lr){4-11}\cmidrule(l){12-13}
        \multirow{8}{*}{\begin{tabular}[c]{@{}c@{}}Swin~\cite{liu2021swin}\end{tabular}}
         & {MSP}~\cite{msp17iclr}                     & prob                                       & {\(91.35\)}                                    & {\(34.96\)}                          & {\(85.21\)}                                    & {\(51.90\)}                             & {\(94.76\)}                                    & {\(23.19\)}                  & {\(78.97\)}                                    & {\(63.70\)}                  & {\(87.57\)}                                    & {\(43.44\)}                  \\
         & {Energy}~\cite{energyood20nips}            & logit                                      & {\(90.93\)}                                    & {\(27.58\)}                          & {\(82.62\)}                                    & {\(51.57\)}                             & {\(95.22\)}                                    & {\(15.47\)}                  & {\(82.29\)}                                    & {\(45.70\)}                  & {\(87.77\)}                                    & {\(35.08\)}                  \\
         & {ODIN}~\cite{odin18iclr}                   & prob+grad                                  & {\(91.38\)}                                    & {\(28.42\)}                          & {\(85.74\)}                                    & {\(44.59\)}                             & {\(94.24\)}                                    & {\(19.65\)}                  & {\(80.62\)}                                    & {\(53.65\)}                  & {\(88.00\)}                                    & {\(36.58\)}                  \\
         & {MaxLogit}~\cite{hendrycks2019scaling}     & logit                                      & {\(91.91\)}                                    & {\(26.79\)}                          & {\(84.67\)}                                    & {\(47.42\)}                             & {\(95.72\)}                                    & {\(15.41\)}                  & {\(81.28\)}                                    & {\(51.50\)}                  & {\(88.40\)}                                    & {\(35.28\)}                  \\
         & {KL Matching}~\cite{hendrycks2019scaling}  & prob                                       & {\(91.92\)}                                    & {\(40.05\)}                          & {\(86.89\)}                                    & {\(52.93\)}                             & {\(94.77\)}                                    & {\(27.62\)}                  & {\(81.91\)}                                    & {\(67.35\)}                  & {\(88.87\)}                                    & {\(46.99\)}                  \\
         & {Residual\(^\dagger\)}                     & feat                                       & \underline{\(94.64\)}                          & {\(32.19\)}                          & \underline{\(91.31\)}                          & {\(43.97\)}                             & \underline{\(98.89\)}                          & \(~~4.81\)                   & \underline{\(86.68\)}                          & {\(68.55\)}                  & \underline{\(92.88\)}                          & {\(37.38\)}                  \\
         & {ReAct}~\cite{sun2021tone}                 & feat                                       & {\(93.58\)}                                    & \boldmath\(23.07\)                   & {\(85.51\)}                                    & \(47.91\)                               & {\(97.51\)}                                    & \(~~9.98\)                   & {\(84.09\)}                                    & \boldmath\(44.50\)           & \(90.17\)                                      & \(31.36\)                    \\
         & {Mahalanobis}~\cite{mahananobis18nips}     & feat+label                                 & \underline{\(94.57\)}                          & {\(33.41\)}                          & \underline{\(89.92\)}                          & {\(49.17\)}                             & \underline{\(98.69\)}                          & \(~~5.43\)                   & \underline{\(85.46\)}                          & {\(73.55\)}                  & \underline{\(92.16\)}                          & {\(40.39\)}                  \\
         & \textbf{ViM (Ours)}                        & feat+logit                                 & \boldmath{\(96.04\)}                           & {\(23.88\)}                          & \boldmath{\(92.34\)}                           & \boldmath{\(38.49\)}                    & \boldmath{\(99.28\)}                           & \boldmath\(~~2.60\)          & \boldmath{\(88.78\)}                           & {\(59.20\)}                  & \boldmath{\(94.11\)}                           & \boldmath{\(31.04\)}         \\
        \cmidrule(r){1-3}\cmidrule(lr){4-11}\cmidrule(l){12-13}
        \multirow{8}{*}{\begin{tabular}[c]{@{}c@{}}DeiT~\cite{pmlr111}\end{tabular}}
         & {MSP}~\cite{msp17iclr}                     & prob                                       & {\(84.04\)}                                    & {\(62.03\)}                          & {\(81.99\)}                                    & {\(64.48\)}                             & {\(88.25\)}                                    & {\(52.00\)}                  & {\(63.65\)}                                    & {\(87.20\)}                  & {\(79.48\)}                                    & {\(66.43\)}                  \\
         & {Energy}~\cite{energyood20nips}            & logit                                      & {\(74.50\)}                                    & {\(67.21\)}                          & {\(77.47\)}                                    & {\(64.77\)}                             & {\(78.63\)}                                    & {\(65.82\)}                  & {\(60.60\)}                                    & {\(82.75\)}                  & {\(72.80\)}                                    & {\(70.14\)}                  \\
         & {ODIN}~\cite{odin18iclr}                   & prob+grad                                  & {\(80.19\)}                                    & \boldmath{\(59.53\)}                 & {\(81.26\)}                                    & \boldmath{\(59.38\)}                    & {\(85.36\)}                                    & {\(51.81\)}                  & {\(61.70\)}                                    & {\(84.95\)}                  & {\(77.13\)}                                    & \boldmath{\(63.92\)}         \\
         & {MaxLogit}~\cite{hendrycks2019scaling}     & logit                                      & {\(80.11\)}                                    & {\(60.83\)}                          & {\(80.45\)}                                    & {\(60.89\)}                             & {\(85.22\)}                                    & {\(52.54\)}                  & {\(61.38\)}                                    & {\(83.70\)}                  & {\(76.79\)}                                    & {\(64.49\)}                  \\
         & {KL Matching}~\cite{hendrycks2019scaling}  & prob                                       & {\(87.49\)}                                    & {\(60.66\)}                          & \boldmath{\(84.89\)}                           & {\(63.47\)}                             & {\(90.54\)}                                    & \boldmath{\(50.47\)}         & {\(71.05\)}                                    & {\(84.60\)}                  & {\(83.49\)}                                    & {\(64.80\)}                  \\
         & {Residual\(^\dagger\)}                     & feat                                       & \underline{\(88.07\)}                          & {\(69.21\)}                          & {\(82.68\)}                                    & {\(77.75\)}                             & \underline{\(91.32\)}                          & {\(58.30\)}                  & \underline{\(74.54\)}                          & {\(91.25\)}                  & \underline{\(84.15\)}                          & {\(74.13\)}                  \\
         & {ReAct}~\cite{sun2021tone}                 & feat                                       & {\(80.29\)}                                    & \(63.11\)                            & {\(80.45\)}                                    & \(63.99\)                               & {\(84.43\)}                                    & \(59.07\)                    & {\(64.32\)}                                    & \boldmath\(81.85\)           & \(77.37\)                                      & \(67.00\)                    \\
         & {Mahalanobis}~\cite{mahananobis18nips}     & feat+label                                 & \underline{\(89.03\)}                          & {\(66.51\)}                          & \underline{\(83.58\)}                          & {\(77.31\)}                             & \underline{\(91.56\)}                          & {\(58.67\)}                  & \underline{\(75.95\)}                          & {\(90.25\)}                  & \underline{\(85.03\)}                          & {\(73.18\)}                  \\
         & \textbf{ViM (Ours)}                        & feat+logit                                 & \boldmath{\(89.13\)}                           & {\(64.58\)}                          & \underline{\(84.42\)}                          & {\(73.02\)}                             & \boldmath{\(92.15\)}                           & {\(52.79\)}                  & \boldmath{\(95.30\)}                           & {\(89.40\)}                  & \boldmath{\(85.25\)}                           & {\(69.95\)}                  \\
        \bottomrule
    \end{tabular}
    \caption{
        OOD detection for ViM and baseline methods on RepVGG, ResNet-50d, Swin Transformer, and DeiT.
        Their pre-trained weights are used.
        The ID dataset is ImageNet-1K, and OOD datasets are OpenImage-O, Texture, iNaturalist, and ImageNet-O.
        Both metrics AUROC and FPR95 are in percentage.
        The best performing item is bolded, and the second and the third places are underlined.
        The proposed ViM has the largest AUROC and the lowest FPR in most cases.
        \(^\dagger\): Residual is defined in Equ. (4).
    }\label{tab:grand_more}
\end{table*}

In Sec. 6.3 we gave the average AUROC and FPR95 for RepVGG, ResNet-50d, Swin Transformer and DeiT.
We provide the detailed AUROC and FPR95 on OpenImage-O, Texture, iNaturalist, and ImageNet-O in \cref{tab:grand_more}.

\section{Details on OpenImage-O (Sec. 5)}

\begin{figure}[t]
    \begin{center}
        \includegraphics[width=0.45\textwidth]{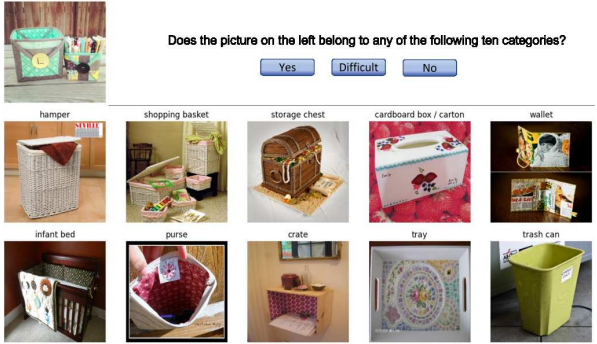}
    \end{center}
    \caption{
        A demonstrative UI for the labelers.
        The image on the left-top corner is the candidate OOD image to be labeled.
        The two rows of images below are from the \(10\) most similar ID classes.
        Labelers choose from \emph{yes/difficult/no} according to these information.
    }\label{fig:labeler}
\end{figure}

\begin{figure*}[t]
    \begin{center}
        \includegraphics[width=0.95\textwidth]{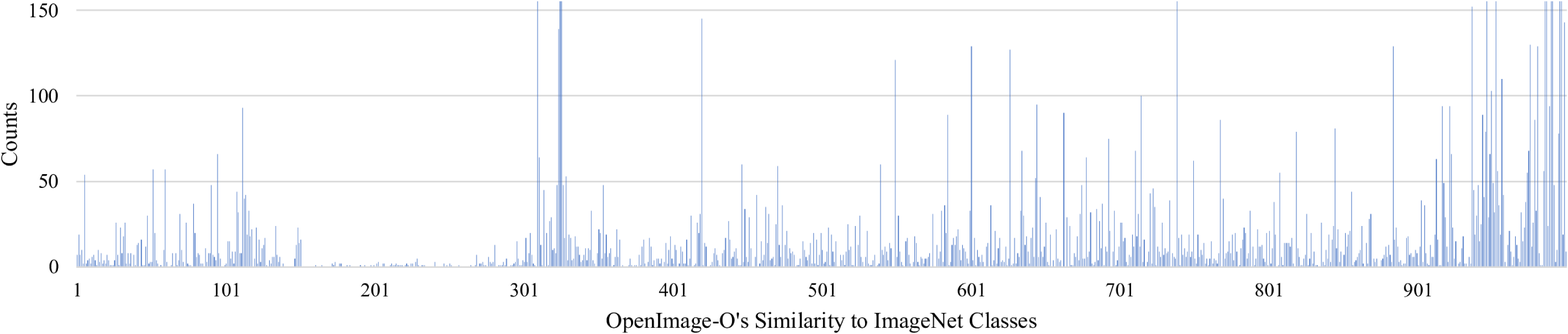}\\
        \includegraphics[width=0.95\textwidth]{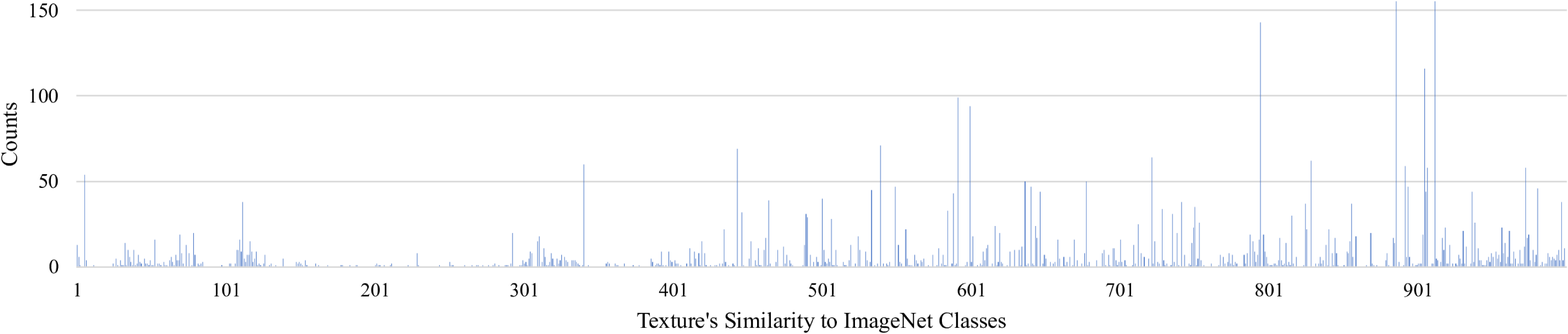}\\
        \includegraphics[width=0.95\textwidth]{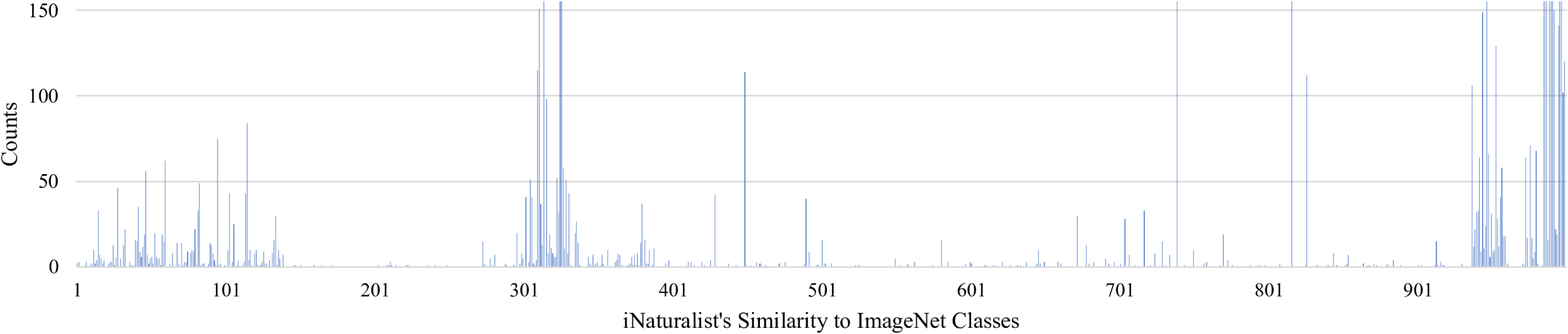}\\
        \includegraphics[width=0.95\textwidth]{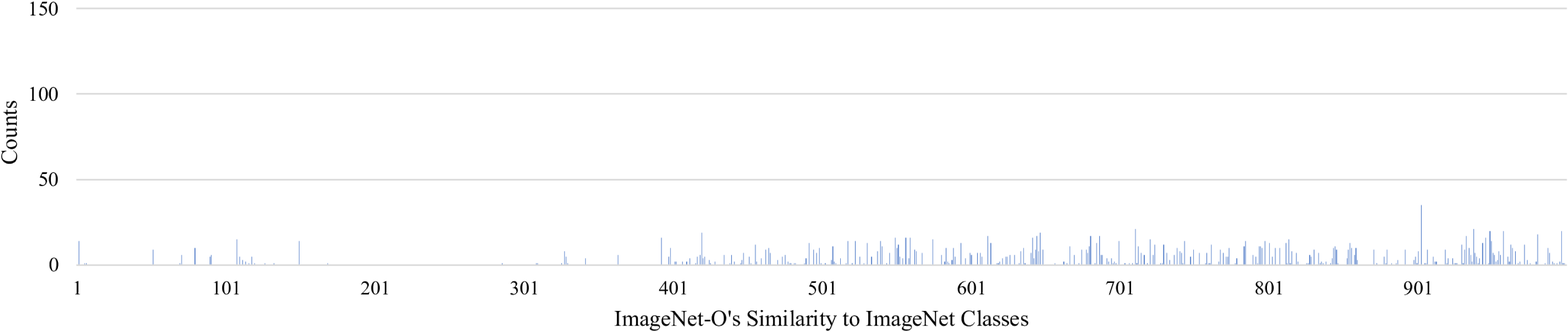}
    \end{center}
    \caption{
        The diversity of four OOD datasets shown by how they look similar to the ImageNet-1K classes.
        We use the BiT model to predict which ID class the image most resembles,
        and count the number of such OOD images for each class.
        Results are shown above.
        Due to space limitation, the \(y\)-axis is clipped at 155.
        Our newly created OpenImage-O has a wider coverage on ImageNet ID classes.
    }\label{fig:data}
\end{figure*}

An illustrative software interface for labelers is shown in \cref{fig:labeler}.
For each candidate OOD image to be labeled,
we find the top 10 classes in ImageNet-1K predicted by a classification model.
Then we gather the most similar images in those top 10 classes by cosine similarity in the feature space.
Next, we patch them as well as their labels with the corresponding OpenImage samples, and let the labelers distinguish whether the OpenImage sample belongs to any of the top 10 categories.
We also set a choice called difficult, so that labelers can put the undistinguishable hard samples into the difficult category.
To reduce annotation noises, each image is labeled twice from different group of labelers.
Then we take the set of OOD images having consensus from the two groups, resulting in an OOD dataset with 17,632 unique images.
In the end, a random inspection process is performed to guarantee the quality of the OOD dataset.

The OpenImage-O follows a natural image distribution as both the source dataset and the labeling process do not involve any filtration based on pre-defined list of labels.
To get a sense of its distribution,
we use the BiT model to find the most similar ID class in ImageNet for each OOD image.
Then the histogram is illustrated in \cref{fig:data}.
It shows that the coverage of OpenImage-O is broader compared to the other three OOD datasets.

\section{Details on Grouping (Sec. 6.5)}

MOS~\cite{huang2021mos} is trained using the officially released code and its default parameter setting.
For all experiments in Sec. 6.5,
the grouping strategy follows the taxonomy grouping defined in~\cite{huang2021mos}.

\begin{table*}[h]
    \centering
    \begin{tabular}{lcccccccccc}
        \toprule
        \multirow{2}{*}{\begin{tabular}[c]{@{}c@{}}\textbf{Method}\end{tabular}} & \multicolumn{2}{c}{\scalebox{.9}[1.0]{\textbf{OpenImage-O}}} & \multicolumn{2}{c}{\textbf{Texture}} & \multicolumn{2}{c}{\scalebox{.95}[1.0]{\textbf{iNaturalist}}} & \multicolumn{2}{c}{\scalebox{.9}[1.0]{\textbf{ImageNet-O}}}                                                                                                                         \\
                                                   & {\small AUROC\(\uparrow\)}                                   & {\small FPR95\(\downarrow\)}         & {\small AUROC\(\uparrow\)}                                    & {\small FPR95\(\downarrow\)}                                & {\small AUROC\(\uparrow\)} & {\small FPR95\(\downarrow\)} & {\small AUROC\(\uparrow\)} & {\small FPR95\(\downarrow\)} \\
        \midrule
        \scalebox{1.0}[1.0]{{MSP} }                & \(92.53\)                                                    & \(34.18\)                            & \(87.10\)                                                     & \(48.55\)                                                   & \(96.11\)                  & \(19.04\)                    & \(81.86\)                  & \(64.85\)                    \\
        \scalebox{1.0}[1.0]{{MaxGroup} }           & {\(92.60\)}                                                  & {\(48.08\)}                          & {\(87.84\)}                                                   & {\(60.08\)}                                                 & {\(95.39\)}                & {\(31.40\)}                  & {\(84.45\)}                & {\(71.90\)}                  \\
        \scalebox{1.0}[1.0]{{ViM} }                & {\(97.61\)}                                                  & {\(12.61\)}                          & \boldmath{\(95.34\)}                                          & \boldmath{\(20.31\)}                                        & \boldmath{\(99.41\)}       & \boldmath{\(~~2.60\)}        & \boldmath{\(92.55\)}       & \boldmath{\(36.75\)}         \\
        \scalebox{1.0}[1.0]{{ViM+Group}}           & \boldmath{\(97.64\)}                                         & \boldmath{\(12.51\)}                 & {\(95.29\)}                                                   & {\(20.41\)}                                                 & {\(99.40\)}                & {\(~~2.70\)}                 & {\(92.50\)}                & {\(37.05\)}                  \\
        \bottomrule
    \end{tabular}
    \caption{
        Comparison of effect of grouping on ViT.
        All numbers are in percentage.
        The grouping is defined in~\cite{huang2021mos} based on taxonomy.
        MaxGroup is the group version of MSP and ViM+Group is the group version of ViM.
    }\label{tab:cluster_vit}
\end{table*}

\begin{table*}[t]
    \centering
    \begin{tabular}{lcccc}
        \toprule
        \textbf{Method} & \textbf{OpenImage-O} & \textbf{Texture} & \textbf{iNaturalist} & \textbf{ImageNet-O} \\
        \midrule
        Residual        & ~~~~~~1.70s          & ~~~~0.56s        & ~~~~~~1.00s          & ~~~~0.19s           \\
        KL Matching     & ~~249.97s            & ~~78.65s         & ~~141.63s            & ~~33.51s            \\
        Mahalanobis     & 2135.13s             & 626.80s          & 1210.82s             & 243.69s             \\
        ViM             & ~~~~~~1.49s          & ~~~~0.51s        & ~~~~~~0.86s          & ~~~~0.18s           \\
        \bottomrule
    \end{tabular}
    \caption{
        Score computation time for four methods on four OOD datasets. We assume that the features have been extracted, so the network forward time is not included.
        The implementation uses numpy and runs on Intel Xeon (Skylake) \@ 23.20GHz CPU.
    }\label{tab:runtime}
\end{table*}

\paragraph{Grouping Results on ViT}
The grouping strategy is less effective for the ViT model,
as seen from results in \cref{tab:cluster_vit}.
Comparing MSP with its group version, MaxGroup, we can see that the improvement on AUROC is very small, while FPRs become even worse.
Examining ViM with its group variant ViM+Group, we can see that their difference is very small, and the original version of ViM is slightly better than ViM+Group.

\section{Details on Baselines (Sec. 6)}

\paragraph{Mahalanobis}

On the BiT model, when including lower level features, the performance of Mahalanobis degrades a lot.
The average AUROC on the four OOD datasets is 56\%, which is much worse than the baseline MSP\@.
Similar results is also found in~\cite[Table 1]{huang2021mos}.
In this paper, we implement the Mahalanobis score using the feature vector before the final classification fc layer, as in~\cite{nearood21arxiv}.
The precision matrix and the class-wise average vector are estimated using 200,000 random training samples.
The ground-truth class label is used during computation.

\paragraph{KL Matching}

We estimate the class-wise average probability using 200,000 random training samples.
Following the practice of~\cite{hendrycks2019scaling}, the predicted class is used instead of ground-truth labels.
We would like to note that the hyperparameter selection for OOD methods should not base on the ID set that is used for computing FPR95 and AUROC (in our case, its the validation set of ImageNet), because once the OOD method overfits the validation set, the evaluation result can be higher than the actual performance.

\paragraph{ReAct}

For ReAct, we use the Energy+ReAct setting, which is the most effective settings in~\cite{sun2021tone}.
In the original paper, they recommended the 90-th percentile of activations estimated on the ID data for the clipping threshold.
However, for BiT and ViT, we found that the rectification percentile \(p=99\) works much better than 90.
So we report results using \(p=99\).

\section{OOD Examples Detected by KL Matching and Residual (Sec. 3)}

\begin{figure*}[t]
    \begin{center}
        \includegraphics[width=0.92\textwidth]{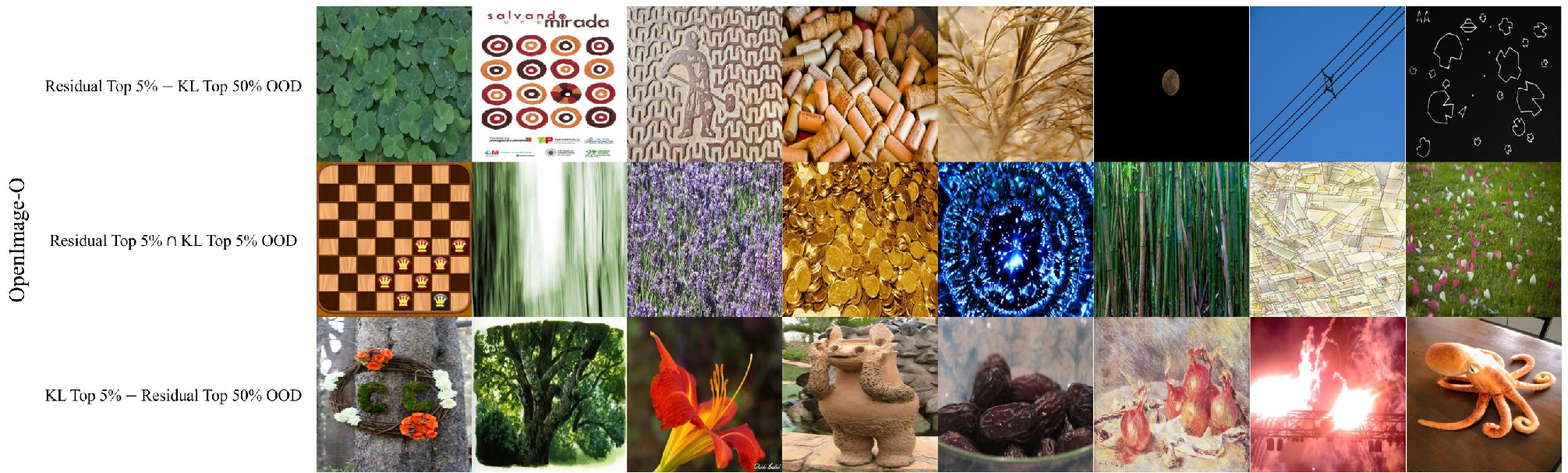}\\
        % \vskip{0.2cm}
        \includegraphics[width=0.92\textwidth]{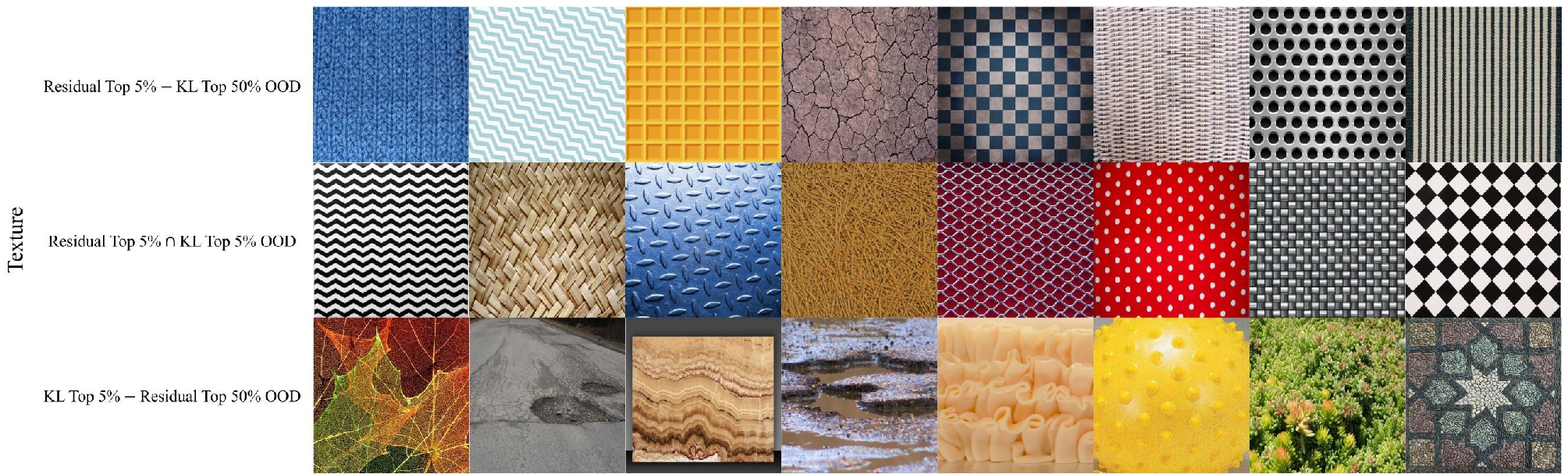}\\
        % \vspace{0.2cm}
        \includegraphics[width=0.92\textwidth]{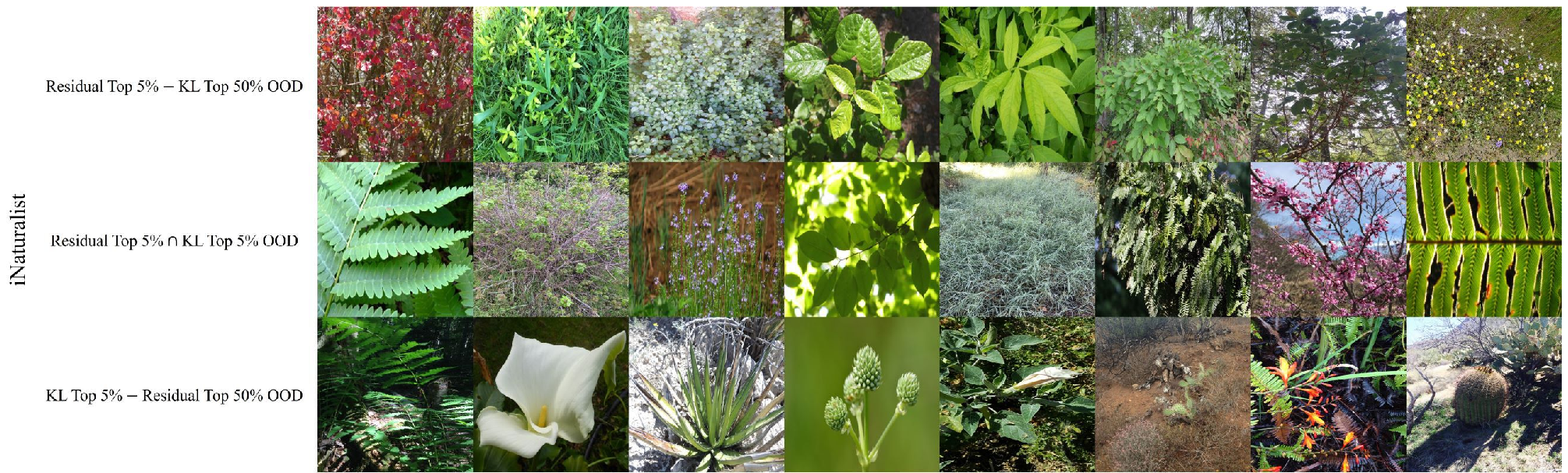}\\
        % \vspace{0.2cm}
        \includegraphics[width=0.92\textwidth]{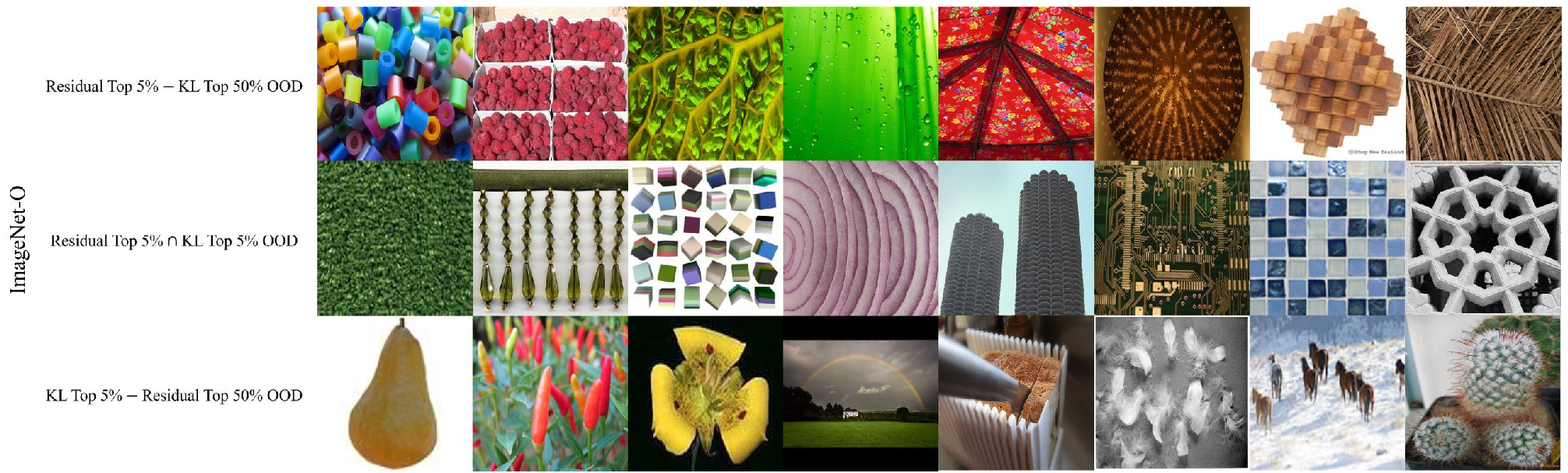}
    \end{center}
    \caption{
        OOD examples detected by Residual and KL Matching.
        There are three rows for each OOD dataset.
        The first row shows images from the top 5\% OODs detected by Residual, with overlapping images in the top 50\% list of KL Matching removed.
        The second row displays images from the intersection of the top 5\% OODs detected by Residual and the top 5\% OODs detected by KL Matching.
        The third row shows images from the top 5\% OODs detected by KL Matching, with overlapping ones in the top 50\% list of Residual removed.
    }\label{fig:diffex}
\end{figure*}

In Sec. 3, we showed that feature-based OOD scores (\eg Residual) and logit/softmax-based OOD scores (\eg KL Matching) have different performances on the Texture OOD dataset.
Here we visualize the OOD examples found by the two methods in \cref{fig:diffex}.

\section{Running Time of Four Methods (Sec. 6.2)}

From Tab. 2 and Tab. 6, it is clear that the four most competitive methods are ViM, Mahalanobis, KL Matching, and Residual.
Our ViM is the fastest among all four methods.
We show their inference time on the four datasets in \cref{tab:runtime}.

{\small
    \bibliographystyle{ieee_fullname}
    \bibliography{bib}
}

\end{document}